\documentclass[sigconf]{acmart}

\copyrightyear{2024}
\acmYear{2024}
\setcopyright{acmlicensed}\acmConference[SIGGRAPH Conference Papers '24]{Special Interest Group on Computer Graphics and Interactive Techniques Conference Conference Papers '24}{July 27-August 1, 2024}{Denver, CO, USA}
\acmBooktitle{Special Interest Group on Computer Graphics and Interactive Techniques Conference Conference Papers '24 (SIGGRAPH Conference Papers '24), July 27-August 1, 2024, Denver, CO, USA}
\acmDOI{10.1145/3641519.3657518}
\acmISBN{979-8-4007-0525-0/24/07}

\usepackage{arydshln}
\usepackage{bbding} 

\begin{document}

\title{MotionCtrl: A Unified and Flexible Motion Controller for Video Generation}

\author{Zhouxia Wang}
\authornote{Works done while as an intern in ARC Lab, Tencent PCG.}
\email{wzhoux@connect.hku.hk}
\orcid{0000-0003-4677-5760}
\affiliation{%
  \institution{S-Lab, Nanyang Technological University}
  \country{Singapore}
}

\author{Ziyang Yuan}
\authornotemark[1]
\email{yuanzy22@mails.tsinghua.edu.cn}
\affiliation{%
  \institution{Tsinghua University}
  \country{China}
}

\author{Xintao Wang}
\authornote{Corresponding author.}
\email{xintao.alpha@gmail.com}
\affiliation{%
  \institution{ARC Lab, Tencent PCG}
  \country{China}
}

\author{Yaowei Li}
\authornotemark[1]
\email{ywl@stu.pku.edu.cn}
\affiliation{%
  \institution{Peking University}
  \country{China}
}

\author{Tianshui Chen}
\email{tianshuichen@gmail.com}
\affiliation{%
  \institution{Guangdong University of Technology}
  \country{China}
}

\author{Menghan Xia}
\email{menghanxyz@gmail.com}
\affiliation{%
  \institution{Tencent AI Lab}
  \country{China}
}

\author{Ping Luo}
\authornotemark[2]
\email{pluo@cs.hku.hk}
\affiliation{%
  \institution{University of Hong Kong}
  \country{China}
}

\author{Ying Shan}
\email{yingsshan@tencent.com}
\affiliation{%
  \institution{ARC Lab, Tencent PCG}
  \country{China}
}

\renewcommand{\shortauthors}{Wang, et al.}
\renewcommand{\shortauthors}{Z. Wang, Z. Yuan, X. Wang, Y. Li, T. Chen, M. Xia, P. Luo, and Y. Shan}

\begin{abstract}
  Motions in a video primarily consist of camera motion, induced by camera movement, and object motion, resulting from object movement. Accurate control of both camera and object motion is essential for video generation. However, existing works either mainly focus on one type of motion or do not clearly distinguish between the two, limiting their control capabilities and diversity. Therefore, this paper presents MotionCtrl, a unified and flexible motion controller for video generation designed to effectively and independently control camera and object motion. The architecture and training strategy of MotionCtrl are carefully devised, taking into account the inherent properties of camera motion, object motion, and imperfect training data. Compared to previous methods, MotionCtrl offers three main advantages: 1) It effectively and independently controls camera motion and object motion, enabling more fine-grained motion control and facilitating flexible and diverse combinations of both types of motion. 2) Its motion conditions are determined by camera poses and trajectories, which are appearance-free and minimally impact the appearance or shape of objects in generated videos. 3) It is a relatively generalizable model that can adapt to a wide array of camera poses and trajectories once trained. Extensive qualitative and quantitative experiments have been conducted to demonstrate the superiority of MotionCtrl over existing methods. Project page: \textcolor{magenta}{ \href{https://wzhouxiff.github.io/projects/MotionCtrl/}{https://wzhouxiff.github.io/projects/MotionCtrl/}}.
\end{abstract}



\begin{teaserfigure}
  \includegraphics[width=\textwidth]{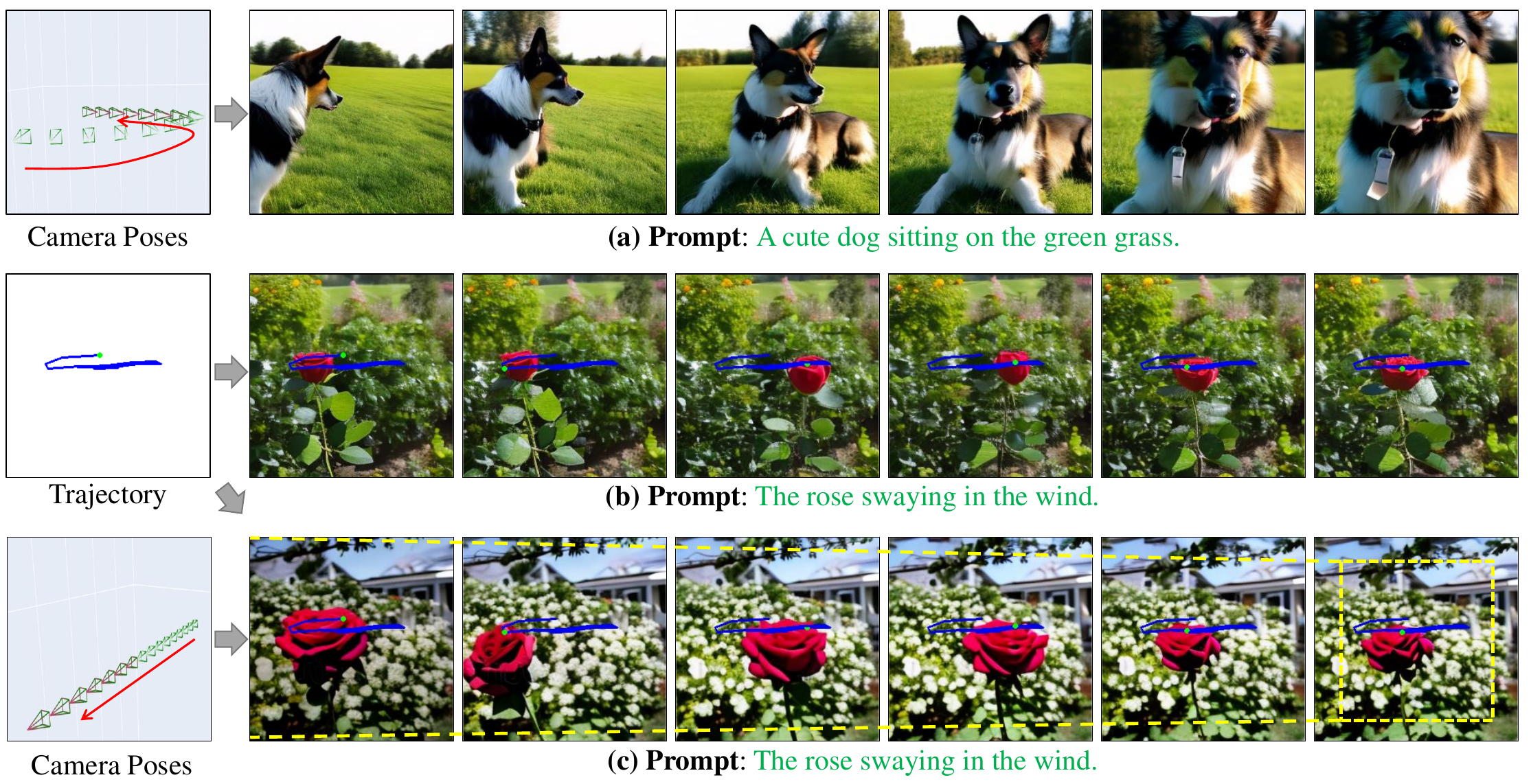}
  \vspace{-0.4cm}
  \caption{\textbf{Control Results of MotionCtrl}. MotionCtrl is capable of controlling both camera motion and object motion in videos produced by a video generation model. It can also simultaneously control both types of motion within the same video. \textbf{We highly encourage readers to check our project page for video results, which cannot be well demonstrated by still images.}}
  \label{fig:teaser}
\end{teaserfigure}

\maketitle

\section{Introduction}
Video generation, such as text-to-video (T2V) generation~\cite{ho2022imagen,singer2022make,zhou2022magicvideo,he2022lvdm,blattmann2023align,chen2023videocrafter1} aims to produce diverse and high-quality videos that conform to given text prompts. Unlike image generation~\cite{t2i1, t2i2, t2i3, t2i4, t2i5, ldm}, which focuses on generating a single image, video generation necessitates the creation of consistent and fluent motion among a sequence of generated images. Consequently, motion control plays a significantly crucial role in video generation, yet it has received limited attention in recent research.

In a video, there are primarily two types of motion: global motion induced by camera movements and local motion resulting from object movements (examples are referred to the zoom out camera poses and swaying rose in Fig.~\ref{fig:teaser} (c)). It should be noted that these two motions will be consistently referred to as \textbf{camera motion} and \textbf{object motion} throughout the paper, respectively. However, most previous works related to motion control in video generation either primarily focus on one of the motions or lack a clear distinction between these two types of motion. For instance, AnimateDiff~\cite{guo2023animatediff}, Gen-2~\cite{esser2023structure}, and PikaLab~\cite{pikalab} mainly execute or trigger camera motion control using independent LoRA~\cite{lora} models or extra camera parameters (such as "-camera zoom in" in PikaLab~\cite{pikalab}). VideoComposer~\cite{wang2023videocomposer} and DragNUWA~\cite{yin2023dragnuwa} implement both camera motion and object motion using the same conditions: motion vector in VideoComposer~\cite{wang2023videocomposer} and trajectory in DragNUWA~\cite{yin2023dragnuwa}. The lack of clear distinction between these two motions prevents these approaches from achieving fine-grained and diverse motion control in video generation.

In this paper, we introduce MotionCtrl, a unified and flexible motion controller for video generation, designed to independently control camera and object motion with a unified model. This approach enables fine-grained motion control in video generation and facilitates flexible and diverse combinations of both motion types.
However, constructing such a unified motion controller presents significant challenges due to the following two factors. First, camera and object motions differ significantly in terms of movement range and pattern. Camera motion refers to the global transformation of the whole scene across the temporal dimension, which is typically represented through a sequence of camera poses over time. In contrast, object motion involves the temporal movement of specific objects within the scene, and it is usually represented as the trajectory of a cluster of pixels associated with the objects. Second, no existing dataset encompasses video clips that are accompanied by a complete set of annotations, including captions, camera poses, and object movement trajectories. Creating such a comprehensive dataset requires a significant amount of effort and resources.

To address the aforementioned challenges, MotionCtrl deploys a delicately designed architecture, training strategy, and curated datasets. MotionCtrl consists of two modules: the Camera Motion Control Module (CMCM) and the Object Motion Control Module (OMCM), each tailored to handle camera motion and object motion characteristics, respectively. Both CMCM and OMCM function as adapter-like modules integrated into existing video generation models. Specifically, CMCM temporally integrates a sequence of camera poses into the video generation model through its temporal transformers, aligning the global motion of the generated video with the provided camera poses. On the other hand, OMCM spatially incorporates information regarding object movement into the convolutional layers of the video generation model, indicating the spatial positioning of objects in each generated frame. Noted that in this study, we utilize VideoCrafter1~\cite{chen2023videocrafter1}, an enhanced version of LVDM~\cite{he2022lvdm}, as the underlying video generation model, which we refer to as LVDM throughout this paper.

Leveraging the delicately designed architecture reliant on a large-scale pre-trained video diffusion model equipped with adapter-like CMCM and OMCM, we can train these modules separately, thereby mitigating the need for a comprehensive dataset containing videos with annotations of captions, camera poses, and object movement trajectories. Consequently, we achieve MotionCtrl with two datasets: one contains annotations of captions and camera poses, and another comprises annotations of captions and object movement trajectories.
Specifically, we introduce the augmented-Realestate10k dataset, originally annotated with camera movement information. We further enhance this dataset by generating captions using Blip2~\cite{blip2}, rendering it suitable for training camera motion control in video generation.
Additionally, we augment videos sourced from WebVid~\cite{bain2021frozen} with object movement trajectories synthesized using the motion segmentation algorithm proposed in ParticleSfM~\cite{zhao2022particlesfm}. Alongside their original annotated captions, the augmented-WebVid dataset becomes conducive to learning object motion control in video generation.
By sequentially and respectively training CMCM and OMCM with these two annotated datasets, our MotionCtrl framework achieves the capability to independently or jointly control camera and object motion within a unified video generation model. This approach enables relatively fine-grained and flexible motion control, empowering users with enhanced control over the generated videos.

Through these delicate designs, MotionCtrl demonstrates superiority over previous methods in three aspects: 1) It independently controls camera and object motion, enabling fine-grained adjustments and a variety of motion combinations, as shown in Fig.~\ref{fig:teaser}. 2) It uses camera poses and trajectories for motion conditions, which do not affect the visual appearance, maintaining the objects' natural look in videos. For instance, our MotionCtrl generates a video with a camera motion that closely reflects the reference video, offering a realistic Eiffel Tower, as seen in Fig.~\ref{fig:sota_camerapose} (b). In contrast, VideoComposer~\cite{wang2023videocomposer} relies on dense motion vectors and mistakenly captures a door's shape of the reference video, resulting in an unnatural Eiffel Tower. 3) MotionCtrl can control a variety of camera movements and trajectories, without the need for fine-tuning each individual camera or object motion.

The main contributions of this work can be summarized as follows:
    (1) We introduce MotionCtrl, a unified and flexible motion controller for video generation, designed to independently or jointly control camera motion and object motion in generated videos, achieving more fine-grained and diverse motion control.
    (2) We carefully tailor the architecture and training strategy of MotionCtrl according to the inherent properties of camera motion, object motion, and imperfect training data, effectively achieving fine-grained motion control in video generation.
    (3) We conduct extensive experiments to demonstrate the superiority of MotionCtrl over previous related methods, both qualitatively and quantitatively.

\section{Related Works}
\textcolor{black}{
Early research in video generation primarily relied on Generative Adversarial Networks (GANs) or Variational Autoencoders (VAEs)~\cite{vondrick2016generating,tulyakov2018mocogan,wang2019few,skorokhodov2022stylegan,wang2019few,saito2017temporal}. However, in recent years, with the remarkable capacity demonstrated by diffusion models~\cite{ho2020denoising,ldm,t2i1} in image generation, video generation research has shifted towards utilizing diffusion models. By further incorporating with text~\cite{ho2022imagen,singer2022make,zhou2022magicvideo,he2022lvdm,blattmann2023align,wang2023videocomposer,guo2023animatediff,chen2023videocrafter1} or image~\cite{yin2023nuwa,blattmann2023stable} guidance, diffusion model can generate high-fidelity videos with specific contents. Particularly, the deployment of diffusion models in latent space~\cite{ldm,he2022lvdm,blattmann2023align} has significantly enhanced the computational efficiency of video generation, leading to a surge in downstream research centered on diffusion models. MotionCtrl, for instance, aims to leverage diffusion models for controlling motion in generated videos.
}

\textcolor{black}{
In the areas of motion control of generated videos, many existing approaches learn motion by referencing specific or a series of template videos\cite{wu2022tune,guo2023animatediff,wu2023lamp,zhao2023motiondirector}. While effective at a specific motion control, these methods typically require training a new model for different templates, which can be limiting.
Some efforts aim to achieve more generalized motion control~\cite{chen2023motion,wang2023videocomposer,yin2023dragnuwa}. For instance, VideoComposer~\cite{wang2023videocomposer} introduces motion control via extra provided motion vectors, and DragNUWA~\cite{yin2023dragnuwa} suggests video generation conditioned on an initial image, provided trajectories, and text prompts. However, the motion control in these methods is relatively broad and fails to fine-grainedly disentangle the camera and object motion within videos.
}

\textcolor{black}{
Different from these works, we propose MotionCtrl, a unified and flexible motion controller that can use either the camera poses and object trajectories or combine these two kinds of guidance to control the motion of generated videos. It enables a more fine-grained and flexible control for video generation.
}

\begin{figure*}[t]
  \centering
   \includegraphics[width=0.98\linewidth]{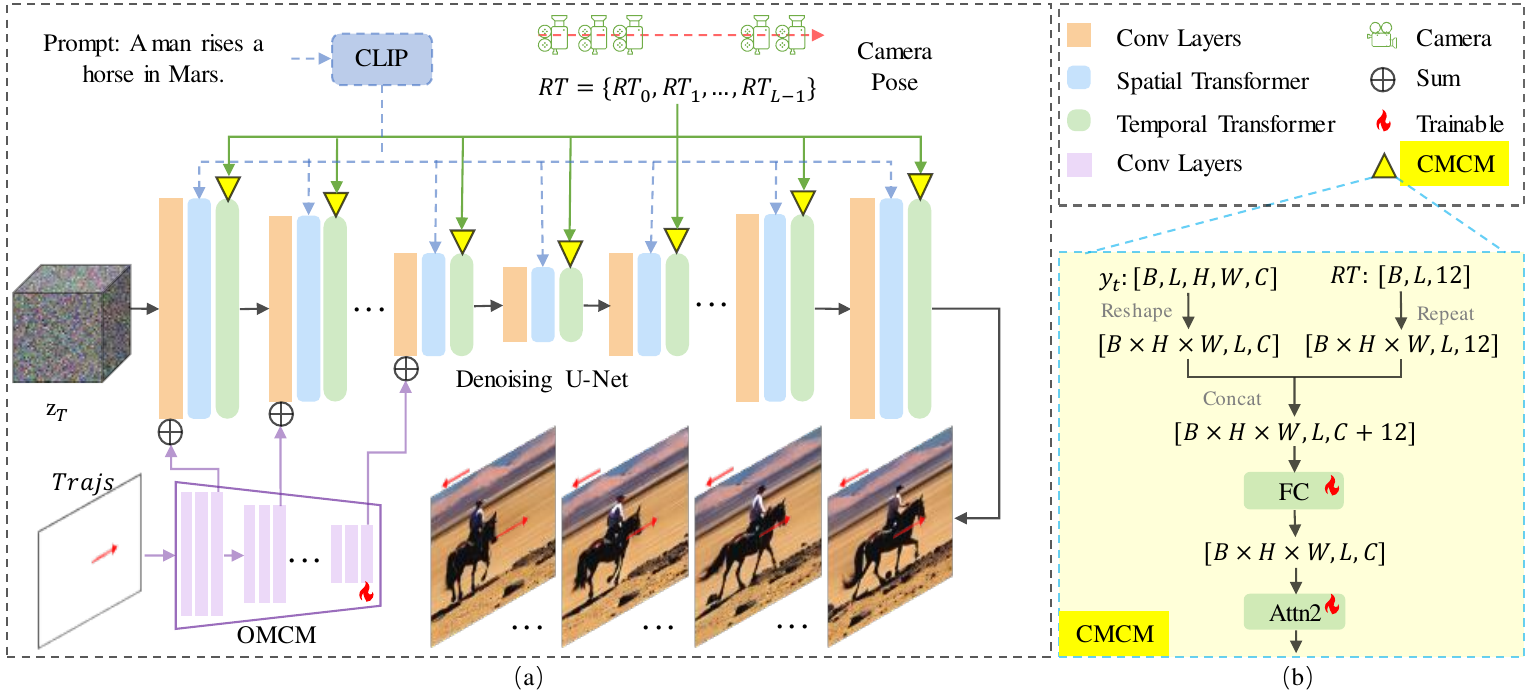}
    \vspace{-0.3cm}
   \caption{\textbf{MotionCtrl Framework}. MotionCtrl extends the Denoising U-Net structure of LVDM with a Camera Motion Control Module (CMCM) and an Object Motion Control Module (OMCM). As illustrated in (b), the CMCM integrates camera pose sequences $RT$ with LVDM's temporal transformers by appending $RT$ to the input of the second self-attention module and applying a tailored and lightweight fully connected layer to extract the camera pose feature for subsequent processing. The OMCM utilizes convolutional layers and downsamplings to derive multi-scale features from $Trajs$, which are spatially incorporated into LVDM's convolutional layers to direct object motion. Further given a text prompt, LVDM generates videos from noise that correspond to the prompt, with background and object movements reflecting the specified camera poses and trajectories. The resulting video demonstrates the horse moving along its trajectory and meanwhile, the background moves left, consistent with the camera's rightward motion.}
   \vspace{-0.27cm}
   \label{fig:framework}
\end{figure*}

\section{Methodology}
\subsection{Preliminary}

The Latent Video Diffusion Model (LVDM)~\cite{he2022lvdm} aims to generate high-quality and diverse videos guided by text prompts. It employs a denoising diffusion model (U-Net~\cite{unet}) in the latent space for space and time efficiency. Consequently, it constructs a lightweight 3D autoencoder, comprising an encoder $\mathcal{E}$ and a decoder $\mathcal{D}$, to encode raw videos into the latent space and reconstruct the denoised latent features back into videos, respectively. Its denoising U-Net (denoted as $\mathcal{\epsilon_\theta}$) is constructed with a sequence of blocks that consist of convolutional layers, spatial transformers, and temporal transformers (shown in Fig.~\ref{fig:framework}). It is optimized using a noise-prediction loss:
\begin{equation}
    \mathcal{L} = \mathbb{E}_{z_0, c, \epsilon\sim\mathcal{N}(0, \mathit{I}), t}\left\lbrack 
\lVert \epsilon - \epsilon_\theta(z_t, t, c) \rVert_2^2 \right\rbrack,
\label{euqation: noise_preditcion_loss}
\end{equation}
where $c$ represents the text prompt, $z_0$ is the latent code obtained using $\mathcal{E}$, $t$ ($t \in [0, T]$) denotes the time step, and $z_t$ is the noisy latent features acquired by weighted addition of Gaussian noise $\epsilon$ to $z_0$ using the following formula:
\begin{equation}
    z_t = \sqrt{\bar{\alpha_t}}z_0 + \sqrt{1-\bar{\alpha_t}}\epsilon,~\bar{\alpha_t} = \prod_{i=1}^t\alpha_t,
\end{equation}
where $\alpha_t$ is used for scheduling the noise strength based on time step $t$.

\subsection{MotionCtrl}

Fig.~\ref{fig:framework} illustrates the framework of MotionCtrl. To achieve disentanglement between camera motion and object motion, and enable independent control of these two types of motion, MotionCtrl comprises two main components: a Camera Motion Control Module (CMCM) and an Object Motion Control Module (OMCM). 
Taking into account the global property of camera motion and the local property of object motion, CMCM interacts with the temporal transformers in LVDM, while OMCM spatially cooperates with the convolutional layers in LVDM. Furthermore, we employ multiple training steps to adapt MotionCtrl to the absence of training data that contains high-quality video clips accompanied by captions, camera poses, and object movement trajectories. In the following subsections, we will provide a detailed description of CMCM and OMCM along with their corresponding training datasets and training strategies.

\subsubsection{Camera Motion Control Module (CMCM)}

The CMCM is a lightweight module constructed with several fully connected layers. Since the camera motions are global transformations between frames in a video, CMCM cooperates with LVDM~\cite{he2022lvdm} via its temporal transformers. Typically, the temporal transformers in LVDM comprise two self-attention modules and facilitate temporal information fusion between video frames. To minimize the impact on LVDM's generative performance, CMCM only involves the second self-attention module in the temporal transformers. Specifically, CMCM takes a sequence of camera poses $RT=\{RT_0, RT_1, \dots, RT_{L-1}\}$ as input. In this paper, the camera pose is represented by its $3 \times 3$ rotation matrix and $3 \times 1$ translation matrix. Consequently, $RT \in \mathbb{R}^{L \times 12}$, where $L$ denotes the length of the generated video. As depicted in Fig.~\ref{fig:framework} (b), $RT$ is extended to $H \times W \times L \times 12$ before being concatenated with the output of the first self-attention module in the temporal transformer ($y_t \in \mathbb{R}^{H\times W\times L\times C}$) along the last dimension, where $H$ and $W$ represent the latent spatial size of the generated video, and $C$ is the number of channels in $y_t$. The concatenated results are then projected back to the size of ${H\times W\times L\times C}$ using a fully connected layer before being fed into the second self-attention module in the temporal transformer.

\subsubsection{Object Motion Control Module (OMCM)}

As depicted in Fig.~\ref{fig:framework}, MotionCtrl controls the object motion of the generated video using trajectories ($Trajs$). Typically, a trajectory is represented as a sequence of spatial positions $\{(x_0, y_0), (x_1, y_1), \dots, (x_{L-1}, y_{L-1})\}$, where $(x_i, y_i), i\in [0, L-1]$ indicates that the trajectory passes through the $i_{th}$ frame at the spatial position $(x, y)$. Particularly, $x \in [0, \hat{W})$ and $y \in [0, \hat{H})$, where $\hat{H}$ and $\hat{W}$ are the height and width of $z_T$, respectively. To explicitly expose the moving speed of the object, we represent $Trajs$ as \\
$\{(0, 0), (u_{(x_1, y_1)}, v_{(x_1, y_1)}), \dots, (u_{(x_{L-1}, y_{L-1})}, v_{(x_{L-1}, y_{L-1})})\}$, where
\begin{equation}
    u_{(x_i, y_i)} = x_{i} - x_{i-1}; v_{(x_i, y_i)} = y_{i} - y_{i-1}; 0 < i < L.
\end{equation}
Denoted that the first frame and the other spatial positions in the subsequent frames that the trajectories do not pass are described as $(0, 0)$. Finally, $Trajs \in \mathbb{R}^{L\times \hat{H} \times \hat{W} \times 2}$.

$Trajs$ is injected into LVDM with OMCM, which is highlighted in the purple block of Fig.~\ref{fig:framework}. OMCM consists of multiple convolutional layers combined with downsampling operations. It extracts multi-scale features from the $Trajs$ and correspondingly adds them to the input of the LVDM's convolutional layers. Drawing inspiration from T2I-Adapter~\cite{mou2024t2i}, the trajectories are only applied to the encoder of the Denoising U-Net to balance the quanlity of the generated video with the ability of object motion control.

\subsubsection{Training Strategy and Data Construction}
\label{sec:training_strategy}
To achieve the control of camera and object motion while generating a video via text prompts, video clips in a training dataset must contain annotations of captions, camera poses, and object movement trajectories. However, a dataset with such comprehensive details is currently unavailable, and assembling one would require considerable effort and resources. To address this challenge, we introduce a multi-step training strategy and train our proposed camera motion control module (CMCM) and object motion control module (OMCM) with distinct augmented datasets tailored to their specific motion control requirements.

\paragraph{Learning the camera motion control module (CMCM)} CMCM only requires a training dataset that contains video clips with annotations of captions and camera poses. Considering that Realestate10K \cite{zhou2018stereo} contains over 60k videos with relatively clean annotations of camera poses, we take it as our training dataset of CMCM. However, employing Realestate10K in MotionCtrl presents two potential challenges: 1) The diversity of scenes is limited in Realestate10K, primarily from real estate videos, potentially compromising the quality of the generated video; and 2) it lacks captions needed for T2V models. 

Regarding the first challenge, we adopt an adapter-like control module (CMCM), with only several new added MLP layers and the second self-attention module of the temporal transformers in LVDM are trainable, and reserving the generation quality of LVDM by freezing most of its parameters. Since the temporal transformers are mainly focus on the learning of global motions, the limited scene diverisity of Realestate10K seldom affects the generation quality of LVDM. This is substantiated by quantitative results presented in Table~\ref{tab:camera_motion}, where the FID~\cite{fid} and FVD~\cite{fvd} metrics indicate that the video quality generated by our MotionCtrl is on par with the LVDM outcomes. 

To address the second challenge, we adopt Blip2~\cite{blip2}, an image captioning algorithm, to generate captions for each video clip in Realestate10K. Details are in the supplementary materials.

\begin{figure}[t]
  \centering
  
   \includegraphics[width=0.9\linewidth]{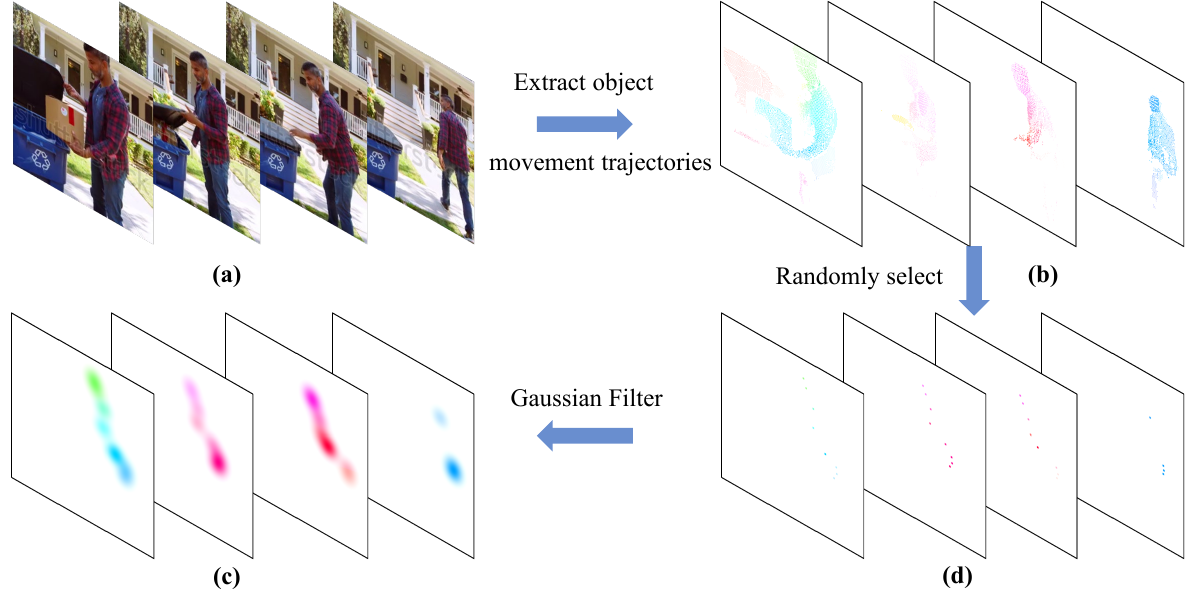}
    \vspace{-0.3cm}
   \caption{\textbf{Trajectories for Object Motion Control}. ParticleSfM~\cite{zhao2022particlesfm} is employed to extract object movement trajectories from video clips, effectively disentangling object motion from camera-induced movement. To circumvent the issues of dense trajectories, which can encode object shapes and are challenging to design at inference, we train the OMCM using sparse trajectories sampled from the dense ones. These sparse trajectories, being too scattered for effective learning, are subsequently refined with a Gaussian filter.}
   \vspace{-18pt}
   \label{fig:particlesfm}
\end{figure}

\paragraph{Learning the object motion control module (OMCM)} OMCM requires a dataset comprising video clips with captions and object movement trajectories, which is currently lacking in the community. To meet the requirement, we utilize ParticleSfM~\cite{zhao2022particlesfm} to synthesize object movement trajectories in WebVid~\cite{bain2021frozen}. WebVid a large-scale video dataset equipped with captions and commonly used in the T2V generation task. Although ParticleSfM is a structure-from-motion system primarily, it incorporates a trajectory-based motion segmentation module utilized for filtering out dynamic trajectories that affect the production of camera trajectories in a dynamic scene. The dynamic trajectories attained by the motion segmentation module exactly fulfill the requirements of our MotionCtrl and we employ this module to synthesize moving object trajectories for about 243,000 videos in WebVid. An example is illustrated in Fig.~\ref{fig:particlesfm} (b), where the trajectories predominantly correspond to a moving person. Synthesis details are in the supplementary.

To circumvent the necessity for users to provide dense trajectories as depicted in Fig.~\ref{fig:particlesfm} (b), which may not be user-friendly, MotionCtrl is required to control the moving objects based on sparse (one or a few) trajectories provided by users. Consequently, our OMCM is trained with $n \in [1, N]$ trajectories (where $N$ represents the maximum number of trajectories for each video) randomly selected from the synthesized dense trajectories (as shown in Fig.~\ref{fig:particlesfm} (c)).
Nevertheless, these selected sparse trajectories tend to be too scattered for effective training. Drawing inspiration from DragNUWA~\cite{yin2023dragnuwa}, we mitigate this issue by applying a Gaussian filter to the sparse trajectories (Fig.~\ref{fig:particlesfm} (d)) and we initially train the OMCM using dense trajectories before fine-tune it using sparse trajectories.

In this training phase, both LVDM and CMCM are well-trained and frozen, with only the OMCM is trained. This strategy guarantees that OMCM adds the object motion control capabilities with a limited dataset while minimally impacting LVDM and CMCM. 
Upon the completion of this training phase, giving both camera poses and object trajectories allows for flexible controlling the camera and object motion in the generated video.

\begin{figure*}[t]
  \centering
  \begin{tabular}{c:c}
       \includegraphics[width=0.43
   \linewidth]{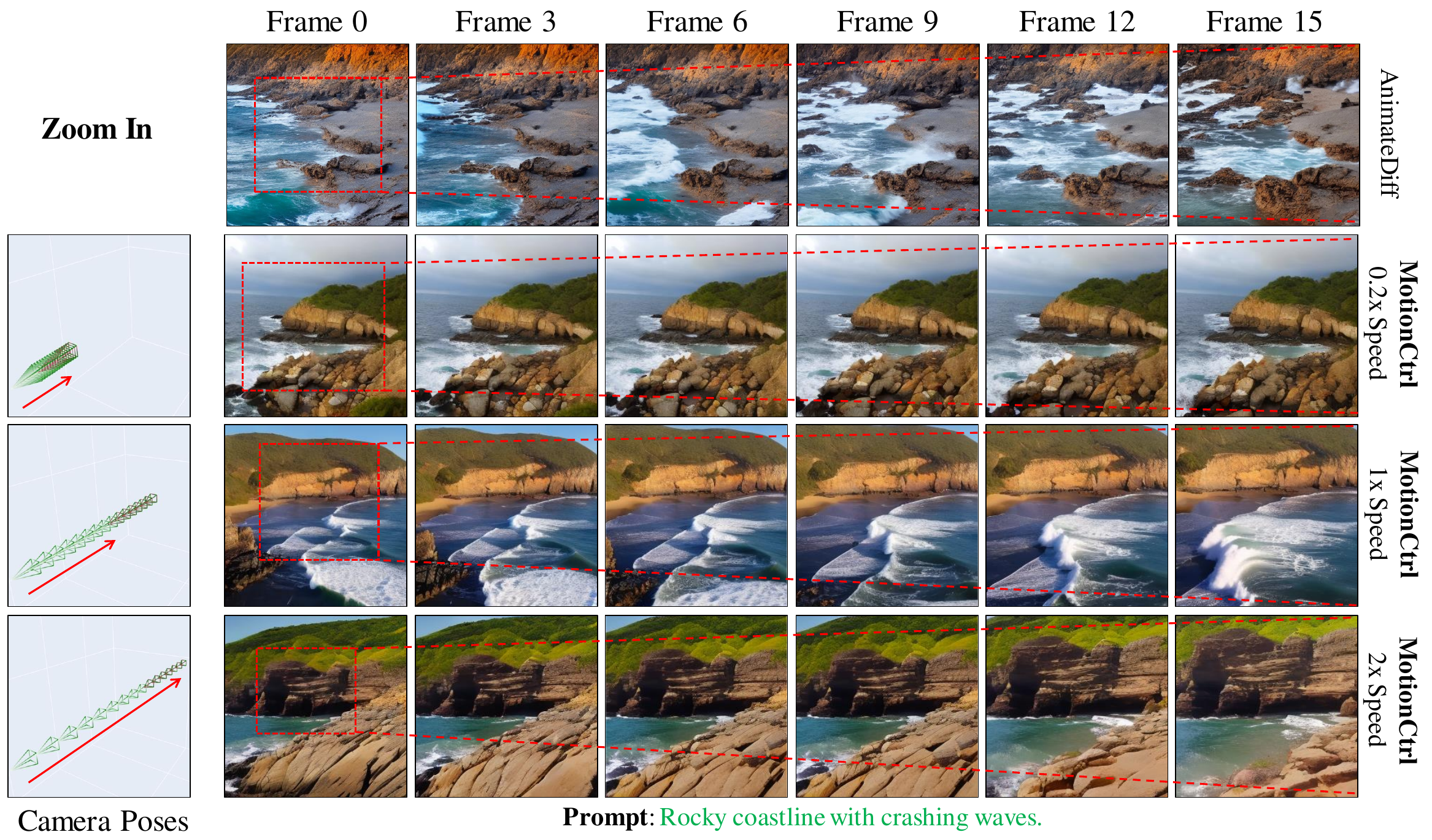}&
   \includegraphics[width=0.53
   \linewidth]{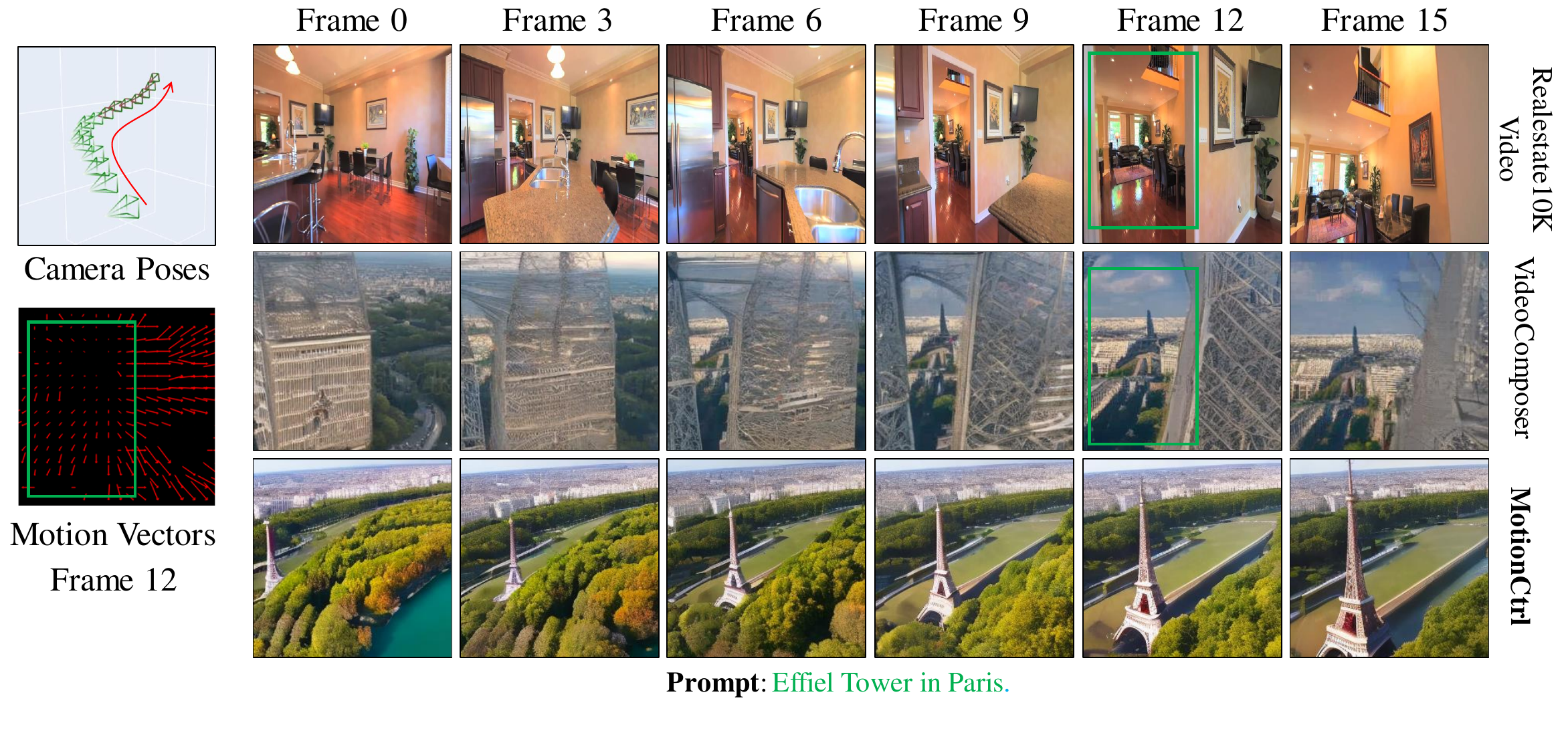}\\
     (a) Camera motion control on basic poses  & (b) Camera motion control on relatively complex poses
  \end{tabular}
  \vspace{-0.3cm}
   \caption{\textbf{Qualitative Comparisons on Camera Motion Control.} (a) Basic Poses: MotionCtrl and AnimateDiff\cite{guo2023animatediff} effectively execute zooms, but MotionCtrl can adjust to varying camera moving speeds. (b) Relatively Complex Poses: VideoComposer\cite{wang2023videocomposer} uses Realestate10K's raw video for motion vectors, capturing unintended shapes like doors, leading to unnatural results (refer to frame 12). MotionCtrl, however, produces a relatively natural video with motion that closely matches the camera poses.}
   \label{fig:sota_camerapose}
\end{figure*}

\section{Experiments}

\subsection{Experiment Settings}

\subsubsection{Implementation Details.}
MotionCtrl is built upon the LVDM framework~\cite{he2022lvdm}/VideoCraft1~\cite{chen2023videocrafter1}, which is trained on 16-frame sequences at a resolution of $256 \times 256$. It can be readily adapted to other video generation models with similar structures, such as AnimateDiff~\cite{guo2023animatediff}, adhering to the settings specific to each model. Additionally, the maximum number of trajectories $N$ is fixed at 8. Both CMCM and OMCM are optimized using the Adam optimizer~\cite{adam} with a batch size of 128 and a learning rate of $1e^{-4}$ across 8 NVIDIA Tesla V100 GPUs. The CMCM typically requires approximately 50,000 iterations to converge. Meanwhile, OMCM undergoes an initial training phase on dense trajectories for 20,000 iterations, followed by fine-tuning with sparse trajectories for an additional 20,000 iterations.

\subsubsection{Evaluation Datasets.}

(1) Camera motion control evaluation dataset encompasses two types of camera poses: basic camera poses (pan left, pan right, pan up, pan down, zoom in, zoom out, anticlockwise rotation, and clockwise rotation) and relatively complex camera poses\footnote{"Complex camera poses" in this work denotes camera movement beyond the basic camera poses. While basic camera poses involve movement in a single straight direction, complex camera poses contain movement in several directions.} obtained from the test set of Realestate10K~\cite{zhou2018stereo} or synthesized using ParticleSfM~\cite{zhao2022particlesfm} on videos from WebVid~\cite{bain2021frozen} and HD-VILA~\cite{xue2022advancing}. 
(2) Object motion control evaluation dataset consists of 283 samples constructed with diverse handcrafted trajectories and prompts.
Further details regarding the construction of the evaluation datasets are provided in the supplementary materials.

\subsubsection{Evaluation Metrics.}
(1) The quality of the generated videos is evaluated using Fréchet Inception Distance (\textbf{FID})\cite{fid}, Fréchet Video Distance (\textbf{FVD})\cite{fvd}, and CLIP Similarity (\textbf{CLIPSIM})~\cite{clip}, which measure the visual quality, temporal coherence, and semantic similarity to the text, respectively. Denoted that the reference videos of FID and FVD are 1000 videos from WebVid~\cite{bain2021frozen}. 
(2) The efficacy of the camera and object motion control is quantified by computing the Euclidean distance between the predicted and ground truth camera poses and object trajectories, respectively. The camera poses and object trajectories of the predicted videos are extracted using ParticleSfM~\cite{zhao2022particlesfm}. We title these two metrics as \textbf{CamMC} and \textbf{ObjMC}, respectively. \textcolor{black}{3) We also conduct a user study for subjective quantitative evaluation, the details of which are provided in the supplementary materials due to space limitations.
}

\begin{figure*}[t]
  \centering
   \includegraphics[width=0.81
   \linewidth]{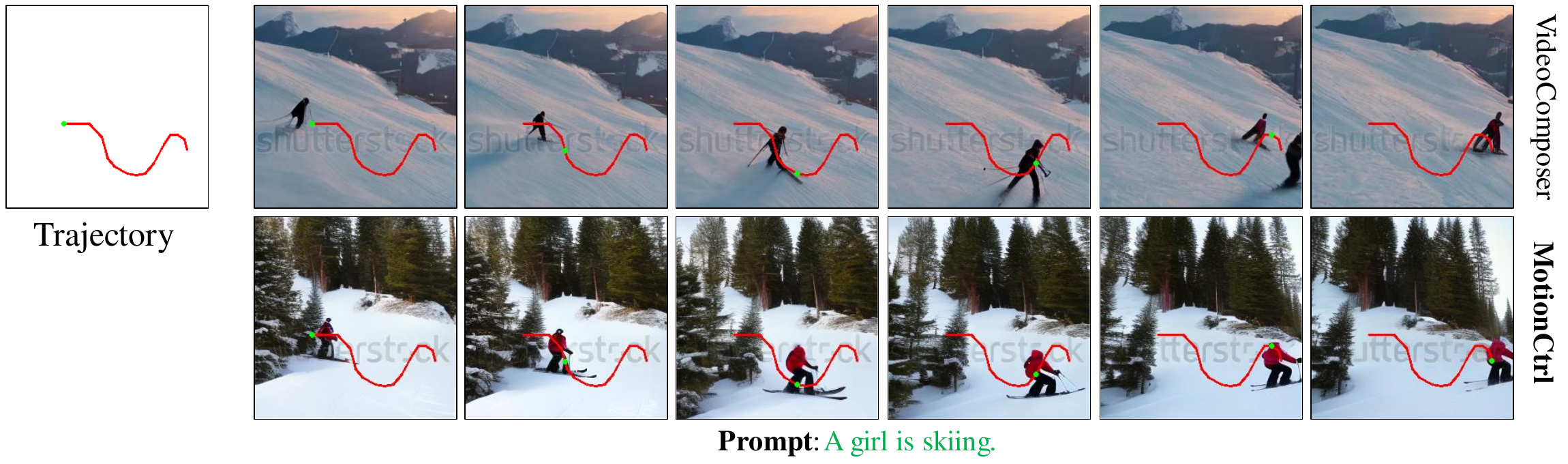}
   \vspace{-0.3cm}
   \caption{\textbf{Qualitative Comparisons on Object Motion Control.} Both VideoComposer and MotionCtrl can generate an object moving along a given trajectory (\textcolor{red}{red curve}), but MotionCtrl more precisely follows it in each frame, as indicated by \textcolor{green}{green points}.}
   \vspace{-8pt}
   \label{fig:sota_traj}
\end{figure*}

\subsection{Comparisons with State-of-the-Art Methods}

To validate the effectiveness of our MotionCtrl in controlling both camera and object motion, we compare it with two leading methods: AnimateDiff~\cite{guo2023animatediff} and VideoComposer~\cite{wang2023videocomposer}. AnimateDiff employs 8 separate LoRA~\cite{lora} models to control 8 basic camera motions in videos, such as panning and zooming, while VideoComposer manipulates video motion using motion vectors without differentiating between camera and object movements. Although DragNUWA~\cite{yin2023dragnuwa} is relevant to our research, its code is not publicly available, precluding a direct comparison. Moreover, DragNUWA only learns motion control with the trajectories extracted from optical flow, which cannot fine-grainedly distinguish the movement between foreground objects and background, limiting its ability to precisely control camera and object motion.

We compare our MotionCtrl with these methods in terms of camera motion and object motion control, and show the capability of our MotionCtrl to flexibly combine the control of camera motion and object motion in video generation. \textit{More comparisons and video comparisons are provided in the supplementary materials.}

\subsubsection{Camera Motion Control.}
We assess camera motion control using basic poses and relatively complex poses. AnimateDiff~\cite{guo2023animatediff} is limited to basic camera poses, while VideoComposer~\cite{wang2023videocomposer} handles complex poses by extracting motion vectors from provided videos. The qualitative results are shown in Fig.~\ref{fig:sota_camerapose}. For basic poses, both MotionCtrl and AnimateDiff can produce videos with forward camera movement, but MotionCtrl can generate camera motion with varying speeds, while AnimateDiff is nonadjustable. Regarding complex poses, where the camera first moves left front and then forward, VideoComposer can mimic the reference video's camera motion using extracted motion vectors. However, the dense motion vectors inadvertently capture object shapes, the door's outline in the reference video (frame 12), resulting in an unnatural-looking Eiffel Tower. MotionCtrl, guided by rotation and translation matrices, generates more natural-looking videos with camera motion close to the reference.

Quantitative results in Table~\ref{tab:sota} show MotionCtrl's superiority over AnimateDiff and VideoComposer for both basic and relatively complex poses, as reflected by the CamMC score. Additionally, MotionCtrl achieves better text similarity and quality metrics, as measured by CLIPSIM, FID, and FVD.

\begin{table}
  \caption{\textcolor{black}{\textbf{Quantitative Comparisons with AnimateDiff~\cite{guo2023animatediff} and VideoComposer~\cite{wang2023videocomposer}.} Our \textbf{MotionCtrl} outperforms competing approaches in both camera and object motion control while also excelling at preserving text similarity and the quality of the video generation.}}
  \vspace{-0.3cm}
  \centering
  \resizebox{\linewidth}{!}{
  \begin{tabular}{c|ccc}
    \toprule
    Method & AnimateDiff & VideoComposer & \textbf{MotionCtrl} \\
    \midrule
    \textbf{CamMC~$\downarrow$} (Basic Poses)     & 0.0548  & -       & \textbf{0.0289} \\
    \textbf{CamMC~$\downarrow$} (Complex Poses)   & -       & 0.0950  & \textbf{0.0735}\\
    \textbf{ObjMC~$\downarrow$}                   & -       & 36.8351 & \textbf{28.877}\\
    \textbf{CLIPSIM~$\uparrow$}                   & 0.2144  & 0.2214  & \textbf{0.2319}\\
    \textbf{FID~$\downarrow$}                     & 157.73  & 130.97  & \textbf{124.09}\\
    \textbf{FVD~$\downarrow$}                     & 1815.88 & 1004.99 & \textbf{852.15}\\ 
    \bottomrule
  \end{tabular}}
  \vspace{-13pt}
  \label{tab:sota}
\end{table}

\subsubsection{Object Motion Control.}
We compare our MotionCtrl with VideoComposer for object motion control, where VideoComposer utilizes motion vectors extracted from trajectories. The qualitative results are shown in Fig.~\ref{fig:sota_traj}. The \textcolor{red}{red curve} illustrates the given trajectory, while the \textcolor{green}{green points} indicate the expected object locations in the corresponding frame. The visual comparison reveals that MotionCtrl can generate objects whose movements are closer to the given trajectories, whereas VideoComposer's results deviate in certain frames, highlighting MotionCtrl's superior object motion control capability. The quantitative results in terms of ObjMC in Table~\ref{tab:sota} also demonstrate that MotionCtrl achieves better object motion control than VideoComposer.

\subsubsection{Combination of Camera Motion and Object Motion.} MotionCtrl can not only control camera and object motion independently within a single video but also perform integrated control of both. As demonstrated in Fig.~\ref{fig:teaser} (b) and (c), when MotionCtrl is applied with only a trajectory, it primarily generates a swaying rose that follows this path. By further introducing zoom-out camera poses, both the rose and the background are animated in accordance with the specified trajectory and camera movements. 

\textcolor{black}{
More results of MotionCtrl can be found in Fig.~\ref{fig:more_results}, supplementary materials, and the demo video.
}

\begin{table}
  \renewcommand{\arraystretch}{1.2}
  \centering
  \caption{\textbf{Ablation of Camera Motion Control.} Our Camera Motion Control Module (CMCM), incorporated with the temporal transformers of  LVDM~\cite{he2022lvdm}, effectively controls camera motion and maintains LVDM's video quality.}
  \vspace{-0.3cm}
  \resizebox{\linewidth}{!}{
  \begin{tabular}{c|cccc}
    \toprule
    Method & \textbf{CamMC~$\downarrow$} & \textbf{CLIPSIM~$\uparrow$}  & \textbf{FID~$\downarrow$} & \textbf{FVD~$\downarrow$} \\
    \midrule
    LVDM~\cite{he2022lvdm}  & 0.9010      & 0.2359 & \textbf{130.62} & 1007.63  \\
    Time Embedding          & 0.0887 & 0.2361 & 132.74 & 1461.36 \\
    Spatial Cross-Attention & 0.0857 & 0.2357 & 153.86 & 1306.78 \\
    Spatial Self-Attention  & 0.0902 & \textbf{0.2384} & 146.37 & 1303.58 \\
    \textbf{Temporal Transformer} & \textbf{0.0289} & 0.2355 & 132.36 & \textbf{1005.24} \\
    \bottomrule
  \end{tabular}}
  \vspace{-12pt}
  \label{tab:camera_motion}
\end{table}

\subsection{Ablation Studies}
\subsubsection{Integrated Position of Camera Motion Control Module (CMCM).}
We test implementing camera motion control by combining camera poses with the time embedding, spatial cross-attention, or spatial self-attention module in LVDM. Although such methods have succeeded in other types of controlling~\cite{ControlNet,mou2024t2i}, such as sketch and depth, they fail to endow camera control capabilities to LVDM, as evidenced by the CamMC scores in Table~\ref{tab:camera_motion} and visualized results in Fig.~\ref{fig:abl_camerapose}. Their CamMC scores are close to the original LVDM.
That is because these components primarily focus on spatial content generation, which is insensitive to the camera motion encoded in camera poses.
Conversely, incorporating CMCM with LVDM's temporal transformers significantly improves camera motion control, as indicated by a lower CamMC score of $0.0289$ in Table~\ref{tab:camera_motion}. Camera motion primarily causes global view transformations over time, and fusing camera poses into LVDM's temporal blocks aligns with this property, enabling effective camera motion control during video generation.

\subsubsection{Dense Trajectories \textit{v.s.} Sparse Trajectories.} 
OMCM is initially trained with dense object movement trajectories extracted via ParticleSfM~\cite{zhao2022particlesfm} and then fine-tune with sparse trajectories.
We evaluate the effectiveness of this approach by comparing it with training OMCM solely on dense or sparse trajectories. Table~\ref{tab:object_motion} and Fig.~\ref{fig:abl_trajs} indicate that training exclusively with dense trajectories yields inferior outcomes, which is attributed to discrepancies between the training and inference phases (sparse trajectories are provided during inference).
Though training solely with sparse trajectories shows improvement over the dense-only approach, it still falls short of the hybrid method, since sparse trajectories alone provide limited information. In contrast, dense trajectories offer richer information that accelerates learning, and subsequent fine-tuning with sparse trajectories allows OMCM to adjust to the sparsity encountered during inference.

\begin{table}
  \renewcommand{\arraystretch}{1.2}
  \centering
  \caption{\textbf{Ablation of Object Motion Control}. The Object Motion Control Module (OMCM), when initially trained on dense object movement trajectories and subsequently fine-tuned with sparse trajectories, outperforms versions trained exclusively on either dense or sparse trajectories.}
  \vspace{-0.4cm}
  \resizebox{\linewidth}{!}{
  \begin{tabular}{c|cccc}
    \toprule
    Method &  \textbf{ObjMC~$\downarrow$} & \textbf{CLIPSIM~$\uparrow$}  & \textbf{FID~$\downarrow$} & \textbf{FVD~$\downarrow$} \\
    \midrule
    Dense  & 54.4114  & 0.2352 & 175.8622 & 2227.87  \\
    Sparse & 34.6937  & 0.2365 & 158.5553 & 2385.39 \\
    \textbf{Dense + Sparse}    & \textbf{25.1198}  & \textbf{0.2342} & \textbf{149.2754} & \textbf{2001.57} \\
    \bottomrule
  \end{tabular}}
  \vspace{-15pt}
  \label{tab:object_motion}
\end{table}

\subsubsection{Training Strategy.} 
Given the limitations of the available training dataset, we propose a multi-step training strategy for MotionCtrl, starting with the CMCM using Realestate10K~\cite{zhou2018stereo}, followed by the OMCM with synthesized object movement trajectories. To thoroughly assess our approach, we experiment with reversing the order and training OMCM before CMCM. This sequence does not impact camera motion control, as OMCM components do not participate in CMCM training. However, it leads to a decrease in object motion control performance since the subsequent training of CMCM adjusts parts of LVDM's temporal transformers, disrupting the object motion control adaptation achieved during OMCM's initial training. Thus, our multi-step strategy, though a compromise due to dataset constraints, is deliberately structured to train CMCM before OMCM, ensuring enhanced performance in both camera and object motion control.

\subsection{Deploy MotionCtrl on AnimateDiff}
We also deploy our MotionCtrl on AnimateDiff~\cite{guo2023animatediff}. Therefore, we can control the motion of the video generated with our adjusted AnimateDiff cooperating with various LoRA~\cite{lora} models in the committee. Visualized results of complex camera motion control and object motion control are in Fig.~\ref{fig:supp_animate_complex} and Fig.~\ref{fig:supp_animate_traj}. More results are in the supplementary materials.

\section{Limitations}

\textcolor{black}{
As an initial exploration into controlling camera and object motion within a unified video generation model, MotionCtrl has demonstrated promising and insightful results. However, controlling the camera and object motion in the same video with both complex camera and complex object trajectories requires a careful design of these trajectories to achieve a natural and harmonious outcome, and the success rate is relatively low. Further research is needed to enhance the accuracy of simultaneously controlling camera and object motion in generated videos.
}

\section{Conclusion}
This paper proposes MotionCtrl, a unified and flexible controller that can independently or combinably control the camera and object motion in a video attained with a video generation model. To achieve this end, MotionCtrl carefully tailors a camera motion control module and object motion control module to adapt to the specific properties of camera motion and object motion and deploys a multi-step training strategy to train these two modules with delicately augmented datasets. Comprehensive experiments, including qualitative and quantitative evaluations, showcase the superiority of our proposed MotionCtrl in both camera and object motion control.

\bibliographystyle{ACM-Reference-Format}
\bibliography{motionctrl_v1}


\begin{thebibliography}{43}


\ifx \showCODEN    \undefined \def \showCODEN     #1{\unskip}     \fi
\ifx \showDOI      \undefined \def \showDOI       #1{#1}\fi
\ifx \showISBNx    \undefined \def \showISBNx     #1{\unskip}     \fi
\ifx \showISBNxiii \undefined \def \showISBNxiii  #1{\unskip}     \fi
\ifx \showISSN     \undefined \def \showISSN      #1{\unskip}     \fi
\ifx \showLCCN     \undefined \def \showLCCN      #1{\unskip}     \fi
\ifx \shownote     \undefined \def \shownote      #1{#1}          \fi
\ifx \showarticletitle \undefined \def \showarticletitle #1{#1}   \fi
\ifx \showURL      \undefined \def \showURL       {\relax}        \fi
\providecommand\bibfield[2]{#2}
\providecommand\bibinfo[2]{#2}
\providecommand\natexlab[1]{#1}
\providecommand\showeprint[2][]{arXiv:#2}

\bibitem[pik({[n.\,d.]})]%
        {pikalab}
 \bibinfo{year}{[n.\,d.]}\natexlab{}.
\newblock \showarticletitle{https://www.pika.art/}.
\newblock


\bibitem[Bain et~al\mbox{.}(2021)]%
        {bain2021frozen}
\bibfield{author}{\bibinfo{person}{Max Bain}, \bibinfo{person}{Arsha Nagrani}, \bibinfo{person}{G{\"u}l Varol}, {and} \bibinfo{person}{Andrew Zisserman}.} \bibinfo{year}{2021}\natexlab{}.
\newblock \showarticletitle{Frozen in time: A joint video and image encoder for end-to-end retrieval}. In \bibinfo{booktitle}{\emph{ICCV}}.
\newblock


\bibitem[Blattmann et~al\mbox{.}(2023a)]%
        {blattmann2023stable}
\bibfield{author}{\bibinfo{person}{Andreas Blattmann}, \bibinfo{person}{Tim Dockhorn}, \bibinfo{person}{Sumith Kulal}, \bibinfo{person}{Daniel Mendelevitch}, \bibinfo{person}{Maciej Kilian}, \bibinfo{person}{Dominik Lorenz}, \bibinfo{person}{Yam Levi}, \bibinfo{person}{Zion English}, \bibinfo{person}{Vikram Voleti}, \bibinfo{person}{Adam Letts}, {et~al\mbox{.}}} \bibinfo{year}{2023}\natexlab{a}.
\newblock \showarticletitle{Stable video diffusion: Scaling latent video diffusion models to large datasets}.
\newblock \bibinfo{journal}{\emph{arXiv preprint arXiv:2311.15127}} (\bibinfo{year}{2023}).
\newblock


\bibitem[Blattmann et~al\mbox{.}(2023b)]%
        {blattmann2023align}
\bibfield{author}{\bibinfo{person}{Andreas Blattmann}, \bibinfo{person}{Robin Rombach}, \bibinfo{person}{Huan Ling}, \bibinfo{person}{Tim Dockhorn}, \bibinfo{person}{Seung~Wook Kim}, \bibinfo{person}{Sanja Fidler}, {and} \bibinfo{person}{Karsten Kreis}.} \bibinfo{year}{2023}\natexlab{b}.
\newblock \showarticletitle{Align your latents: High-resolution video synthesis with latent diffusion models}. In \bibinfo{booktitle}{\emph{CVPR}}.
\newblock


\bibitem[Chen et~al\mbox{.}(2023b)]%
        {chen2023videocrafter1}
\bibfield{author}{\bibinfo{person}{Haoxin Chen}, \bibinfo{person}{Menghan Xia}, \bibinfo{person}{Yingqing He}, \bibinfo{person}{Yong Zhang}, \bibinfo{person}{Xiaodong Cun}, \bibinfo{person}{Shaoshu Yang}, \bibinfo{person}{Jinbo Xing}, \bibinfo{person}{Yaofang Liu}, \bibinfo{person}{Qifeng Chen}, \bibinfo{person}{Xintao Wang}, {et~al\mbox{.}}} \bibinfo{year}{2023}\natexlab{b}.
\newblock \showarticletitle{Videocrafter1: Open diffusion models for high-quality video generation}.
\newblock \bibinfo{journal}{\emph{arXiv preprint arXiv:2310.19512}} (\bibinfo{year}{2023}).
\newblock


\bibitem[Chen et~al\mbox{.}(2023a)]%
        {chen2023motion}
\bibfield{author}{\bibinfo{person}{Tsai-Shien Chen}, \bibinfo{person}{Chieh~Hubert Lin}, \bibinfo{person}{Hung-Yu Tseng}, \bibinfo{person}{Tsung-Yi Lin}, {and} \bibinfo{person}{Ming-Hsuan Yang}.} \bibinfo{year}{2023}\natexlab{a}.
\newblock \showarticletitle{Motion-conditioned diffusion model for controllable video synthesis}.
\newblock \bibinfo{journal}{\emph{arXiv preprint arXiv:2304.14404}} (\bibinfo{year}{2023}).
\newblock


\bibitem[Ding et~al\mbox{.}(2021)]%
        {t2i3}
\bibfield{author}{\bibinfo{person}{Ming Ding}, \bibinfo{person}{Zhuoyi Yang}, \bibinfo{person}{Wenyi Hong}, \bibinfo{person}{Wendi Zheng}, \bibinfo{person}{Chang Zhou}, \bibinfo{person}{Da Yin}, \bibinfo{person}{Junyang Lin}, \bibinfo{person}{Xu Zou}, \bibinfo{person}{Zhou Shao}, \bibinfo{person}{Hongxia Yang}, {et~al\mbox{.}}} \bibinfo{year}{2021}\natexlab{}.
\newblock \showarticletitle{Cogview: Mastering text-to-image generation via transformers}.
\newblock \bibinfo{journal}{\emph{NeurIPS}} (\bibinfo{year}{2021}).
\newblock


\bibitem[Esser et~al\mbox{.}(2023)]%
        {esser2023structure}
\bibfield{author}{\bibinfo{person}{Patrick Esser}, \bibinfo{person}{Johnathan Chiu}, \bibinfo{person}{Parmida Atighehchian}, \bibinfo{person}{Jonathan Granskog}, {and} \bibinfo{person}{Anastasis Germanidis}.} \bibinfo{year}{2023}\natexlab{}.
\newblock \showarticletitle{Structure and content-guided video synthesis with diffusion models}. In \bibinfo{booktitle}{\emph{ICCV}}.
\newblock


\bibitem[Guo et~al\mbox{.}(2023)]%
        {guo2023animatediff}
\bibfield{author}{\bibinfo{person}{Yuwei Guo}, \bibinfo{person}{Ceyuan Yang}, \bibinfo{person}{Anyi Rao}, \bibinfo{person}{Yaohui Wang}, \bibinfo{person}{Yu Qiao}, \bibinfo{person}{Dahua Lin}, {and} \bibinfo{person}{Bo Dai}.} \bibinfo{year}{2023}\natexlab{}.
\newblock \showarticletitle{AnimateDiff: Animate Your Personalized Text-to-Image Diffusion Models without Specific Tuning}.
\newblock \bibinfo{journal}{\emph{arXiv preprint arXiv:2307.04725}} (\bibinfo{year}{2023}).
\newblock


\bibitem[He et~al\mbox{.}(2022)]%
        {he2022lvdm}
\bibfield{author}{\bibinfo{person}{Yingqing He}, \bibinfo{person}{Tianyu Yang}, \bibinfo{person}{Yong Zhang}, \bibinfo{person}{Ying Shan}, {and} \bibinfo{person}{Qifeng Chen}.} \bibinfo{year}{2022}\natexlab{}.
\newblock \showarticletitle{Latent video diffusion models for high-fidelity long video generation}.
\newblock \bibinfo{journal}{\emph{arXiv preprint arXiv:2211.13221}} (\bibinfo{year}{2022}).
\newblock


\bibitem[Ho et~al\mbox{.}(2022)]%
        {ho2022imagen}
\bibfield{author}{\bibinfo{person}{Jonathan Ho}, \bibinfo{person}{William Chan}, \bibinfo{person}{Chitwan Saharia}, \bibinfo{person}{Jay Whang}, \bibinfo{person}{Ruiqi Gao}, \bibinfo{person}{Alexey Gritsenko}, \bibinfo{person}{Diederik~P Kingma}, \bibinfo{person}{Ben Poole}, \bibinfo{person}{Mohammad Norouzi}, \bibinfo{person}{David~J Fleet}, {et~al\mbox{.}}} \bibinfo{year}{2022}\natexlab{}.
\newblock \showarticletitle{Imagen video: High definition video generation with diffusion models}.
\newblock \bibinfo{journal}{\emph{arXiv preprint arXiv:2210.02303}} (\bibinfo{year}{2022}).
\newblock


\bibitem[Ho et~al\mbox{.}(2020)]%
        {ho2020denoising}
\bibfield{author}{\bibinfo{person}{Jonathan Ho}, \bibinfo{person}{Ajay Jain}, {and} \bibinfo{person}{Pieter Abbeel}.} \bibinfo{year}{2020}\natexlab{}.
\newblock \showarticletitle{Denoising diffusion probabilistic models}.
\newblock \bibinfo{journal}{\emph{NeurIPS}} (\bibinfo{year}{2020}).
\newblock


\bibitem[Hu et~al\mbox{.}(2021)]%
        {lora}
\bibfield{author}{\bibinfo{person}{Edward~J Hu}, \bibinfo{person}{Yelong Shen}, \bibinfo{person}{Phillip Wallis}, \bibinfo{person}{Zeyuan Allen-Zhu}, \bibinfo{person}{Yuanzhi Li}, \bibinfo{person}{Shean Wang}, \bibinfo{person}{Lu Wang}, {and} \bibinfo{person}{Weizhu Chen}.} \bibinfo{year}{2021}\natexlab{}.
\newblock \showarticletitle{Lora: Low-rank adaptation of large language models}.
\newblock \bibinfo{journal}{\emph{arXiv preprint arXiv:2106.09685}} (\bibinfo{year}{2021}).
\newblock


\bibitem[Huang et~al\mbox{.}(2024)]%
        {huang2023vbench}
\bibfield{author}{\bibinfo{person}{Ziqi Huang}, \bibinfo{person}{Yinan He}, \bibinfo{person}{Jiashuo Yu}, \bibinfo{person}{Fan Zhang}, \bibinfo{person}{Chenyang Si}, \bibinfo{person}{Yuming Jiang}, \bibinfo{person}{Yuanhan Zhang}, \bibinfo{person}{Tianxing Wu}, \bibinfo{person}{Qingyang Jin}, \bibinfo{person}{Nattapol Chanpaisit}, {et~al\mbox{.}}} \bibinfo{year}{2024}\natexlab{}.
\newblock \showarticletitle{Vbench: Comprehensive benchmark suite for video generative models}.
\newblock  (\bibinfo{year}{2024}).
\newblock


\bibitem[Kingma and Ba(2014)]%
        {adam}
\bibfield{author}{\bibinfo{person}{Diederik~P Kingma} {and} \bibinfo{person}{Jimmy Ba}.} \bibinfo{year}{2014}\natexlab{}.
\newblock \showarticletitle{Adam: A method for stochastic optimization}.
\newblock \bibinfo{journal}{\emph{arXiv preprint arXiv:1412.6980}} (\bibinfo{year}{2014}).
\newblock


\bibitem[Li et~al\mbox{.}(2023)]%
        {blip2}
\bibfield{author}{\bibinfo{person}{Junnan Li}, \bibinfo{person}{Dongxu Li}, \bibinfo{person}{Silvio Savarese}, {and} \bibinfo{person}{Steven Hoi}.} \bibinfo{year}{2023}\natexlab{}.
\newblock \showarticletitle{Blip-2: Bootstrapping language-image pre-training with frozen image encoders and large language models}. In \bibinfo{booktitle}{\emph{ICML}}.
\newblock


\bibitem[Mou et~al\mbox{.}(2024)]%
        {mou2024t2i}
\bibfield{author}{\bibinfo{person}{Chong Mou}, \bibinfo{person}{Xintao Wang}, \bibinfo{person}{Liangbin Xie}, \bibinfo{person}{Yanze Wu}, \bibinfo{person}{Jian Zhang}, \bibinfo{person}{Zhongang Qi}, {and} \bibinfo{person}{Ying Shan}.} \bibinfo{year}{2024}\natexlab{}.
\newblock \showarticletitle{T2i-adapter: Learning adapters to dig out more controllable ability for text-to-image diffusion models}. In \bibinfo{booktitle}{\emph{AAAI}}.
\newblock


\bibitem[Radford et~al\mbox{.}(2021)]%
        {clip}
\bibfield{author}{\bibinfo{person}{Alec Radford}, \bibinfo{person}{Jong~Wook Kim}, \bibinfo{person}{Chris Hallacy}, \bibinfo{person}{Aditya Ramesh}, \bibinfo{person}{Gabriel Goh}, \bibinfo{person}{Sandhini Agarwal}, \bibinfo{person}{Girish Sastry}, \bibinfo{person}{Amanda Askell}, \bibinfo{person}{Pamela Mishkin}, \bibinfo{person}{Jack Clark}, {et~al\mbox{.}}} \bibinfo{year}{2021}\natexlab{}.
\newblock \showarticletitle{Learning transferable visual models from natural language supervision}. In \bibinfo{booktitle}{\emph{ICML}}.
\newblock


\bibitem[Ramesh et~al\mbox{.}(2022)]%
        {t2i5}
\bibfield{author}{\bibinfo{person}{Aditya Ramesh}, \bibinfo{person}{Prafulla Dhariwal}, \bibinfo{person}{Alex Nichol}, \bibinfo{person}{Casey Chu}, {and} \bibinfo{person}{Mark Chen}.} \bibinfo{year}{2022}\natexlab{}.
\newblock \showarticletitle{Hierarchical text-conditional image generation with clip latents}.
\newblock \bibinfo{journal}{\emph{arXiv preprint arXiv:2204.06125}} (\bibinfo{year}{2022}).
\newblock


\bibitem[Ramesh et~al\mbox{.}(2021)]%
        {t2i2}
\bibfield{author}{\bibinfo{person}{Aditya Ramesh}, \bibinfo{person}{Mikhail Pavlov}, \bibinfo{person}{Gabriel Goh}, \bibinfo{person}{Scott Gray}, \bibinfo{person}{Chelsea Voss}, \bibinfo{person}{Alec Radford}, \bibinfo{person}{Mark Chen}, {and} \bibinfo{person}{Ilya Sutskever}.} \bibinfo{year}{2021}\natexlab{}.
\newblock \showarticletitle{Zero-shot text-to-image generation}. In \bibinfo{booktitle}{\emph{ICML}}.
\newblock


\bibitem[Rombach et~al\mbox{.}(2022)]%
        {ldm}
\bibfield{author}{\bibinfo{person}{Robin Rombach}, \bibinfo{person}{Andreas Blattmann}, \bibinfo{person}{Dominik Lorenz}, \bibinfo{person}{Patrick Esser}, {and} \bibinfo{person}{Bj{\"o}rn Ommer}.} \bibinfo{year}{2022}\natexlab{}.
\newblock \showarticletitle{High-resolution image synthesis with latent diffusion models}. In \bibinfo{booktitle}{\emph{CVPR}}.
\newblock


\bibitem[Ronneberger et~al\mbox{.}(2015)]%
        {unet}
\bibfield{author}{\bibinfo{person}{Olaf Ronneberger}, \bibinfo{person}{Philipp Fischer}, {and} \bibinfo{person}{Thomas Brox}.} \bibinfo{year}{2015}\natexlab{}.
\newblock \showarticletitle{U-net: Convolutional networks for biomedical image segmentation}. In \bibinfo{booktitle}{\emph{MICCAI}}.
\newblock


\bibitem[Saharia et~al\mbox{.}(2022)]%
        {t2i1}
\bibfield{author}{\bibinfo{person}{Chitwan Saharia}, \bibinfo{person}{William Chan}, \bibinfo{person}{Saurabh Saxena}, \bibinfo{person}{Lala Li}, \bibinfo{person}{Jay Whang}, \bibinfo{person}{Emily~L Denton}, \bibinfo{person}{Kamyar Ghasemipour}, \bibinfo{person}{Raphael Gontijo~Lopes}, \bibinfo{person}{Burcu Karagol~Ayan}, \bibinfo{person}{Tim Salimans}, {et~al\mbox{.}}} \bibinfo{year}{2022}\natexlab{}.
\newblock \showarticletitle{Photorealistic text-to-image diffusion models with deep language understanding}.
\newblock \bibinfo{journal}{\emph{NeuIPS}} (\bibinfo{year}{2022}).
\newblock


\bibitem[Saito et~al\mbox{.}(2017)]%
        {saito2017temporal}
\bibfield{author}{\bibinfo{person}{Masaki Saito}, \bibinfo{person}{Eiichi Matsumoto}, {and} \bibinfo{person}{Shunta Saito}.} \bibinfo{year}{2017}\natexlab{}.
\newblock \showarticletitle{Temporal generative adversarial nets with singular value clipping}. In \bibinfo{booktitle}{\emph{ICCV}}.
\newblock


\bibitem[Seitzer(2020)]%
        {fid}
\bibfield{author}{\bibinfo{person}{Maximilian Seitzer}.} \bibinfo{year}{2020}\natexlab{}.
\newblock \bibinfo{title}{{pytorch-fid: FID Score for PyTorch}}.
\newblock \bibinfo{howpublished}{\url{https://github.com/mseitzer/pytorch-fid}}.
\newblock


\bibitem[Singer et~al\mbox{.}(2022)]%
        {singer2022make}
\bibfield{author}{\bibinfo{person}{Uriel Singer}, \bibinfo{person}{Adam Polyak}, \bibinfo{person}{Thomas Hayes}, \bibinfo{person}{Xi Yin}, \bibinfo{person}{Jie An}, \bibinfo{person}{Songyang Zhang}, \bibinfo{person}{Qiyuan Hu}, \bibinfo{person}{Harry Yang}, \bibinfo{person}{Oron Ashual}, \bibinfo{person}{Oran Gafni}, {et~al\mbox{.}}} \bibinfo{year}{2022}\natexlab{}.
\newblock \showarticletitle{Make-a-video: Text-to-video generation without text-video data}.
\newblock \bibinfo{journal}{\emph{arXiv preprint arXiv:2209.14792}} (\bibinfo{year}{2022}).
\newblock


\bibitem[Skorokhodov et~al\mbox{.}(2022)]%
        {skorokhodov2022stylegan}
\bibfield{author}{\bibinfo{person}{Ivan Skorokhodov}, \bibinfo{person}{Sergey Tulyakov}, {and} \bibinfo{person}{Mohamed Elhoseiny}.} \bibinfo{year}{2022}\natexlab{}.
\newblock \showarticletitle{Stylegan-v: A continuous video generator with the price, image quality and perks of stylegan2}. In \bibinfo{booktitle}{\emph{CVPR}}.
\newblock


\bibitem[Tulyakov et~al\mbox{.}(2018)]%
        {tulyakov2018mocogan}
\bibfield{author}{\bibinfo{person}{Sergey Tulyakov}, \bibinfo{person}{Ming-Yu Liu}, \bibinfo{person}{Xiaodong Yang}, {and} \bibinfo{person}{Jan Kautz}.} \bibinfo{year}{2018}\natexlab{}.
\newblock \showarticletitle{Mocogan: Decomposing motion and content for video generation}. In \bibinfo{booktitle}{\emph{CVPR}}.
\newblock


\bibitem[Unterthiner et~al\mbox{.}(2018)]%
        {fvd}
\bibfield{author}{\bibinfo{person}{Thomas Unterthiner}, \bibinfo{person}{Sjoerd Van~Steenkiste}, \bibinfo{person}{Karol Kurach}, \bibinfo{person}{Raphael Marinier}, \bibinfo{person}{Marcin Michalski}, {and} \bibinfo{person}{Sylvain Gelly}.} \bibinfo{year}{2018}\natexlab{}.
\newblock \showarticletitle{Towards accurate generative models of video: A new metric \& challenges}.
\newblock \bibinfo{journal}{\emph{arXiv preprint arXiv:1812.01717}} (\bibinfo{year}{2018}).
\newblock


\bibitem[Vondrick et~al\mbox{.}(2016)]%
        {vondrick2016generating}
\bibfield{author}{\bibinfo{person}{Carl Vondrick}, \bibinfo{person}{Hamed Pirsiavash}, {and} \bibinfo{person}{Antonio Torralba}.} \bibinfo{year}{2016}\natexlab{}.
\newblock \showarticletitle{Generating videos with scene dynamics}.
\newblock \bibinfo{journal}{\emph{NeuIPS}} (\bibinfo{year}{2016}).
\newblock


\bibitem[Wang et~al\mbox{.}(2019)]%
        {wang2019few}
\bibfield{author}{\bibinfo{person}{Ting-Chun Wang}, \bibinfo{person}{Ming-Yu Liu}, \bibinfo{person}{Andrew Tao}, \bibinfo{person}{Guilin Liu}, \bibinfo{person}{Jan Kautz}, {and} \bibinfo{person}{Bryan Catanzaro}.} \bibinfo{year}{2019}\natexlab{}.
\newblock \showarticletitle{Few-shot video-to-video synthesis}.
\newblock \bibinfo{journal}{\emph{arXiv preprint arXiv:1910.12713}} (\bibinfo{year}{2019}).
\newblock


\bibitem[Wang et~al\mbox{.}(2023)]%
        {wang2023videocomposer}
\bibfield{author}{\bibinfo{person}{Xiang Wang}, \bibinfo{person}{Hangjie Yuan}, \bibinfo{person}{Shiwei Zhang}, \bibinfo{person}{Dayou Chen}, \bibinfo{person}{Jiuniu Wang}, \bibinfo{person}{Yingya Zhang}, \bibinfo{person}{Yujun Shen}, \bibinfo{person}{Deli Zhao}, {and} \bibinfo{person}{Jingren Zhou}.} \bibinfo{year}{2023}\natexlab{}.
\newblock \showarticletitle{VideoComposer: Compositional Video Synthesis with Motion Controllability}.
\newblock \bibinfo{journal}{\emph{arXiv preprint arXiv:2306.02018}} (\bibinfo{year}{2023}).
\newblock


\bibitem[Wu et~al\mbox{.}(2023b)]%
        {wu2022tune}
\bibfield{author}{\bibinfo{person}{Jay~Zhangjie Wu}, \bibinfo{person}{Yixiao Ge}, \bibinfo{person}{Xintao Wang}, \bibinfo{person}{Stan~Weixian Lei}, \bibinfo{person}{Yuchao Gu}, \bibinfo{person}{Yufei Shi}, \bibinfo{person}{Wynne Hsu}, \bibinfo{person}{Ying Shan}, \bibinfo{person}{Xiaohu Qie}, {and} \bibinfo{person}{Mike~Zheng Shou}.} \bibinfo{year}{2023}\natexlab{b}.
\newblock \showarticletitle{Tune-a-video: One-shot tuning of image diffusion models for text-to-video generation}. In \bibinfo{booktitle}{\emph{ICCV}}.
\newblock


\bibitem[Wu et~al\mbox{.}(2023a)]%
        {wu2023lamp}
\bibfield{author}{\bibinfo{person}{Ruiqi Wu}, \bibinfo{person}{Liangyu Chen}, \bibinfo{person}{Tong Yang}, \bibinfo{person}{Chunle Guo}, \bibinfo{person}{Chongyi Li}, {and} \bibinfo{person}{Xiangyu Zhang}.} \bibinfo{year}{2023}\natexlab{a}.
\newblock \showarticletitle{LAMP: Learn A Motion Pattern for Few-Shot-Based Video Generation}.
\newblock \bibinfo{journal}{\emph{arXiv preprint arXiv:2310.10769}} (\bibinfo{year}{2023}).
\newblock


\bibitem[Xue et~al\mbox{.}(2022)]%
        {xue2022advancing}
\bibfield{author}{\bibinfo{person}{Hongwei Xue}, \bibinfo{person}{Tiankai Hang}, \bibinfo{person}{Yanhong Zeng}, \bibinfo{person}{Yuchong Sun}, \bibinfo{person}{Bei Liu}, \bibinfo{person}{Huan Yang}, \bibinfo{person}{Jianlong Fu}, {and} \bibinfo{person}{Baining Guo}.} \bibinfo{year}{2022}\natexlab{}.
\newblock \showarticletitle{Advancing high-resolution video-language representation with large-scale video transcriptions}. In \bibinfo{booktitle}{\emph{CVPR}}.
\newblock


\bibitem[Yin et~al\mbox{.}(2023a)]%
        {yin2023dragnuwa}
\bibfield{author}{\bibinfo{person}{Shengming Yin}, \bibinfo{person}{Chenfei Wu}, \bibinfo{person}{Jian Liang}, \bibinfo{person}{Jie Shi}, \bibinfo{person}{Houqiang Li}, \bibinfo{person}{Gong Ming}, {and} \bibinfo{person}{Nan Duan}.} \bibinfo{year}{2023}\natexlab{a}.
\newblock \showarticletitle{Dragnuwa: Fine-grained control in video generation by integrating text, image, and trajectory}.
\newblock \bibinfo{journal}{\emph{arXiv preprint arXiv:2308.08089}} (\bibinfo{year}{2023}).
\newblock


\bibitem[Yin et~al\mbox{.}(2023b)]%
        {yin2023nuwa}
\bibfield{author}{\bibinfo{person}{Shengming Yin}, \bibinfo{person}{Chenfei Wu}, \bibinfo{person}{Huan Yang}, \bibinfo{person}{Jianfeng Wang}, \bibinfo{person}{Xiaodong Wang}, \bibinfo{person}{Minheng Ni}, \bibinfo{person}{Zhengyuan Yang}, \bibinfo{person}{Linjie Li}, \bibinfo{person}{Shuguang Liu}, \bibinfo{person}{Fan Yang}, {et~al\mbox{.}}} \bibinfo{year}{2023}\natexlab{b}.
\newblock \showarticletitle{Nuwa-xl: Diffusion over diffusion for extremely long video generation}.
\newblock \bibinfo{journal}{\emph{arXiv preprint arXiv:2303.12346}} (\bibinfo{year}{2023}).
\newblock


\bibitem[Zhang et~al\mbox{.}(2023)]%
        {ControlNet}
\bibfield{author}{\bibinfo{person}{Lvmin Zhang}, \bibinfo{person}{Anyi Rao}, {and} \bibinfo{person}{Maneesh Agrawala}.} \bibinfo{year}{2023}\natexlab{}.
\newblock \showarticletitle{Adding conditional control to text-to-image diffusion models}. In \bibinfo{booktitle}{\emph{ICCV}}.
\newblock


\bibitem[Zhao et~al\mbox{.}(2023)]%
        {zhao2023motiondirector}
\bibfield{author}{\bibinfo{person}{Rui Zhao}, \bibinfo{person}{Yuchao Gu}, \bibinfo{person}{Jay~Zhangjie Wu}, \bibinfo{person}{David~Junhao Zhang}, \bibinfo{person}{Jiawei Liu}, \bibinfo{person}{Weijia Wu}, \bibinfo{person}{Jussi Keppo}, {and} \bibinfo{person}{Mike~Zheng Shou}.} \bibinfo{year}{2023}\natexlab{}.
\newblock \showarticletitle{MotionDirector: Motion Customization of Text-to-Video Diffusion Models}.
\newblock \bibinfo{journal}{\emph{arXiv preprint arXiv:2310.08465}} (\bibinfo{year}{2023}).
\newblock


\bibitem[Zhao et~al\mbox{.}(2022)]%
        {zhao2022particlesfm}
\bibfield{author}{\bibinfo{person}{Wang Zhao}, \bibinfo{person}{Shaohui Liu}, \bibinfo{person}{Hengkai Guo}, \bibinfo{person}{Wenping Wang}, {and} \bibinfo{person}{Yong-Jin Liu}.} \bibinfo{year}{2022}\natexlab{}.
\newblock \showarticletitle{Particlesfm: Exploiting dense point trajectories for localizing moving cameras in the wild}. In \bibinfo{booktitle}{\emph{ECCV}}.
\newblock


\bibitem[Zhou et~al\mbox{.}(2022)]%
        {zhou2022magicvideo}
\bibfield{author}{\bibinfo{person}{Daquan Zhou}, \bibinfo{person}{Weimin Wang}, \bibinfo{person}{Hanshu Yan}, \bibinfo{person}{Weiwei Lv}, \bibinfo{person}{Yizhe Zhu}, {and} \bibinfo{person}{Jiashi Feng}.} \bibinfo{year}{2022}\natexlab{}.
\newblock \showarticletitle{Magicvideo: Efficient video generation with latent diffusion models}.
\newblock \bibinfo{journal}{\emph{arXiv preprint arXiv:2211.11018}} (\bibinfo{year}{2022}).
\newblock


\bibitem[Zhou et~al\mbox{.}(2018)]%
        {zhou2018stereo}
\bibfield{author}{\bibinfo{person}{Tinghui Zhou}, \bibinfo{person}{Richard Tucker}, \bibinfo{person}{John Flynn}, \bibinfo{person}{Graham Fyffe}, {and} \bibinfo{person}{Noah Snavely}.} \bibinfo{year}{2018}\natexlab{}.
\newblock \showarticletitle{Stereo magnification: Learning view synthesis using multiplane images}.
\newblock \bibinfo{journal}{\emph{arXiv preprint arXiv:1805.09817}} (\bibinfo{year}{2018}).
\newblock


\bibitem[Zhou et~al\mbox{.}(2021)]%
        {t2i4}
\bibfield{author}{\bibinfo{person}{Yufan Zhou}, \bibinfo{person}{Ruiyi Zhang}, \bibinfo{person}{Changyou Chen}, \bibinfo{person}{Chunyuan Li}, \bibinfo{person}{Chris Tensmeyer}, \bibinfo{person}{Tong Yu}, \bibinfo{person}{Jiuxiang Gu}, \bibinfo{person}{Jinhui Xu}, {and} \bibinfo{person}{Tong Sun}.} \bibinfo{year}{2021}\natexlab{}.
\newblock \showarticletitle{Lafite: Towards language-free training for text-to-image generation}.
\newblock \bibinfo{journal}{\emph{arXiv preprint arXiv:2111.13792}} (\bibinfo{year}{2021}).
\newblock


\end{thebibliography}

\appendix

\begin{figure*}[t]
\hsize=\textwidth
\centering
\includegraphics[width=0.8\textwidth]{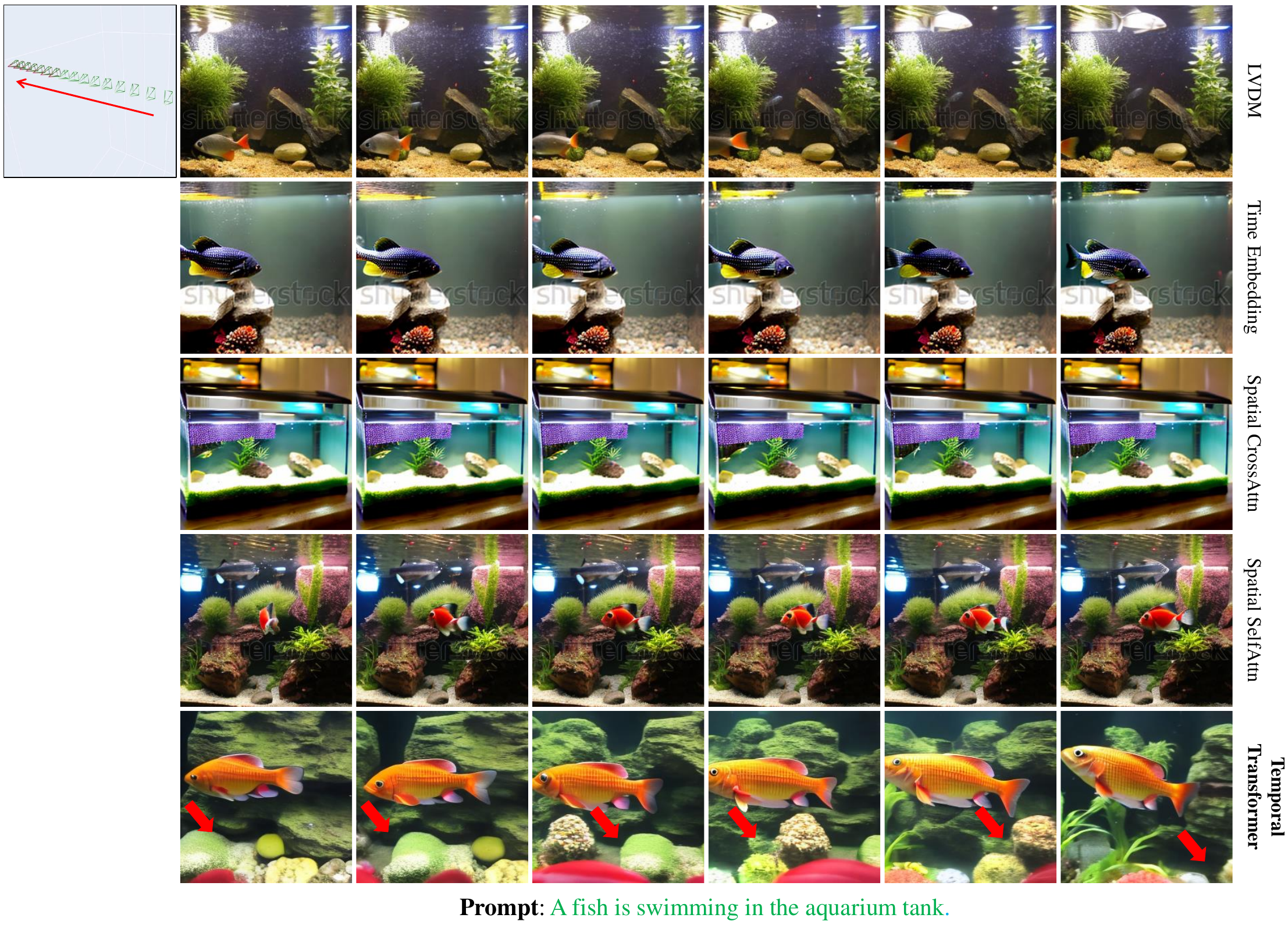} 
\caption{\textbf{The qualitative results of ablation study regarding the integrated position of the Camera Motion Control Module (CMCM) with LVDM~\cite{he2022lvdm}.} Integrating CMCM of MotionCtrl with the temporal transformers in LVDM significantly improves camera motion control compared to other setups.}
\label{fig:abl_camerapose} 
\end{figure*}

\begin{figure*}[t]
\hsize=\textwidth
\centering
\includegraphics[width=0.8\textwidth]{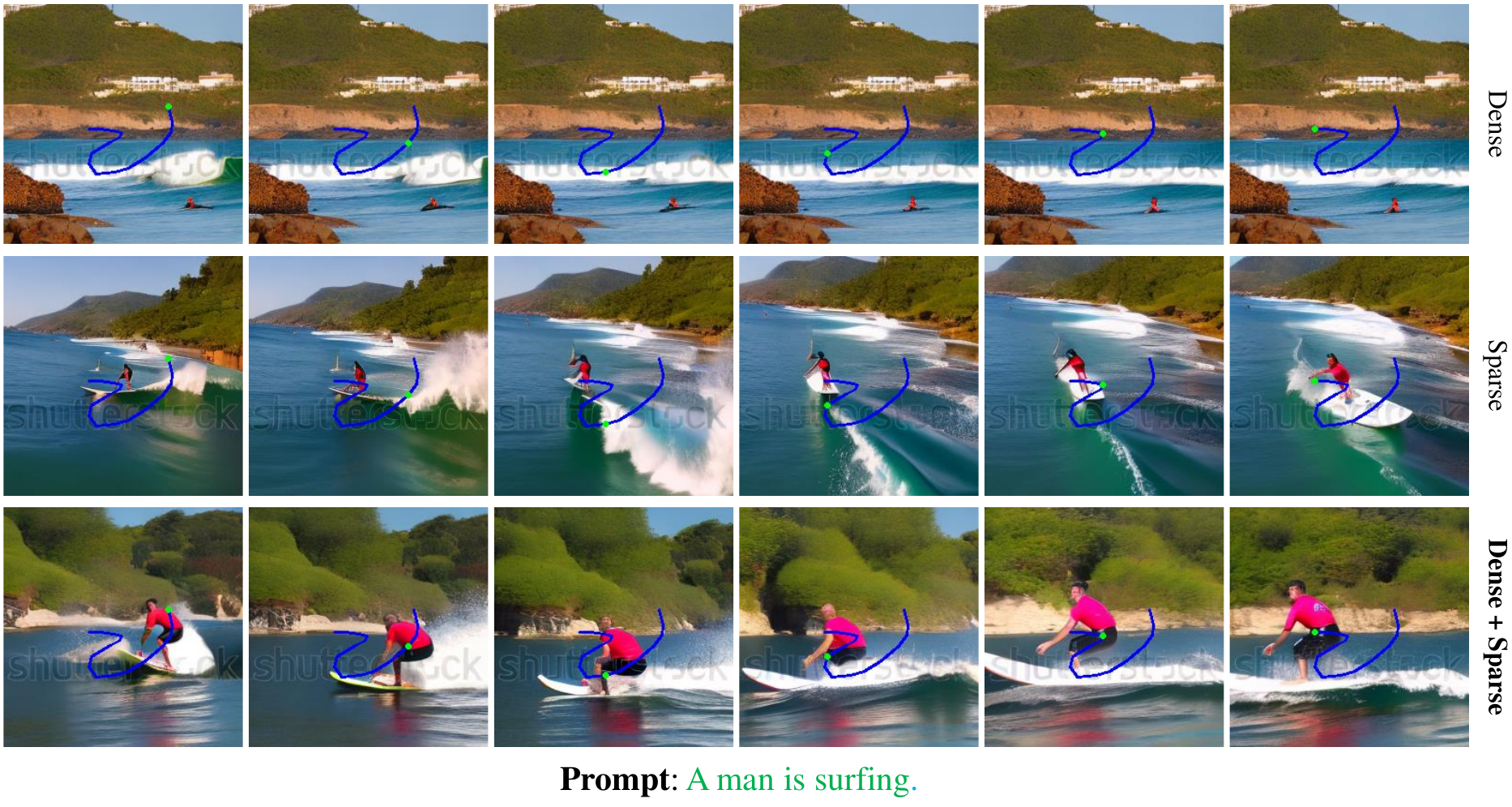} 
\caption{\textbf{The qualitative results of ablation study "Dense Trajectories v.s. Sparse Trajectories"}. The model trained with dense trajectories fails to control the object motion in the generated video. Conversely, the model trained on dense trajectories, followed by fine-tuning on sparse trajectories, exhibits superior precision in object motion control compared to the model trained solely on sparse trajectories.}
\label{fig:abl_trajs} 
\end{figure*}

\begin{figure*}[t]
\hsize=\textwidth
\centering
\includegraphics[width=0.73\textwidth]{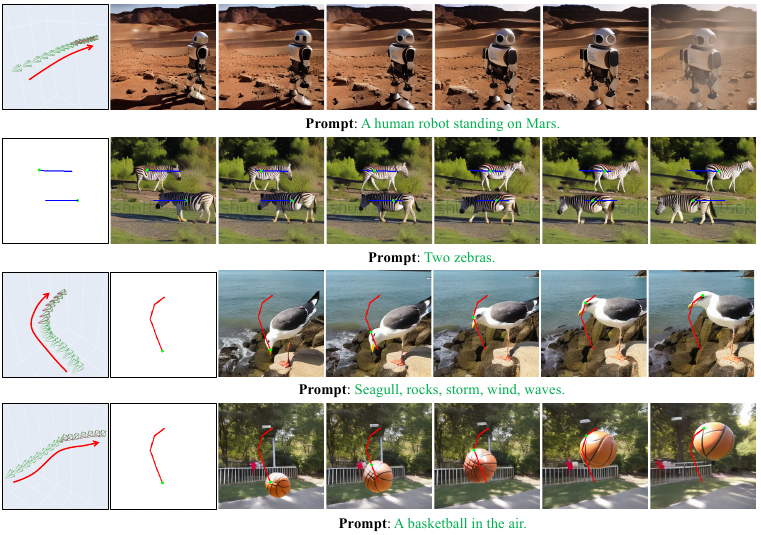} 
\vspace{-0.3cm}
\caption{\textcolor{black}{More results of MotionCtrl include those controlled by camera poses or object trajectories independently, as well as those controlled with camera poses and object trajectories simultaneously.}}
\label{fig:more_results}
\end{figure*}

\begin{figure*}[t]
\hsize=\textwidth
\centering
\includegraphics[width=0.64\textwidth]{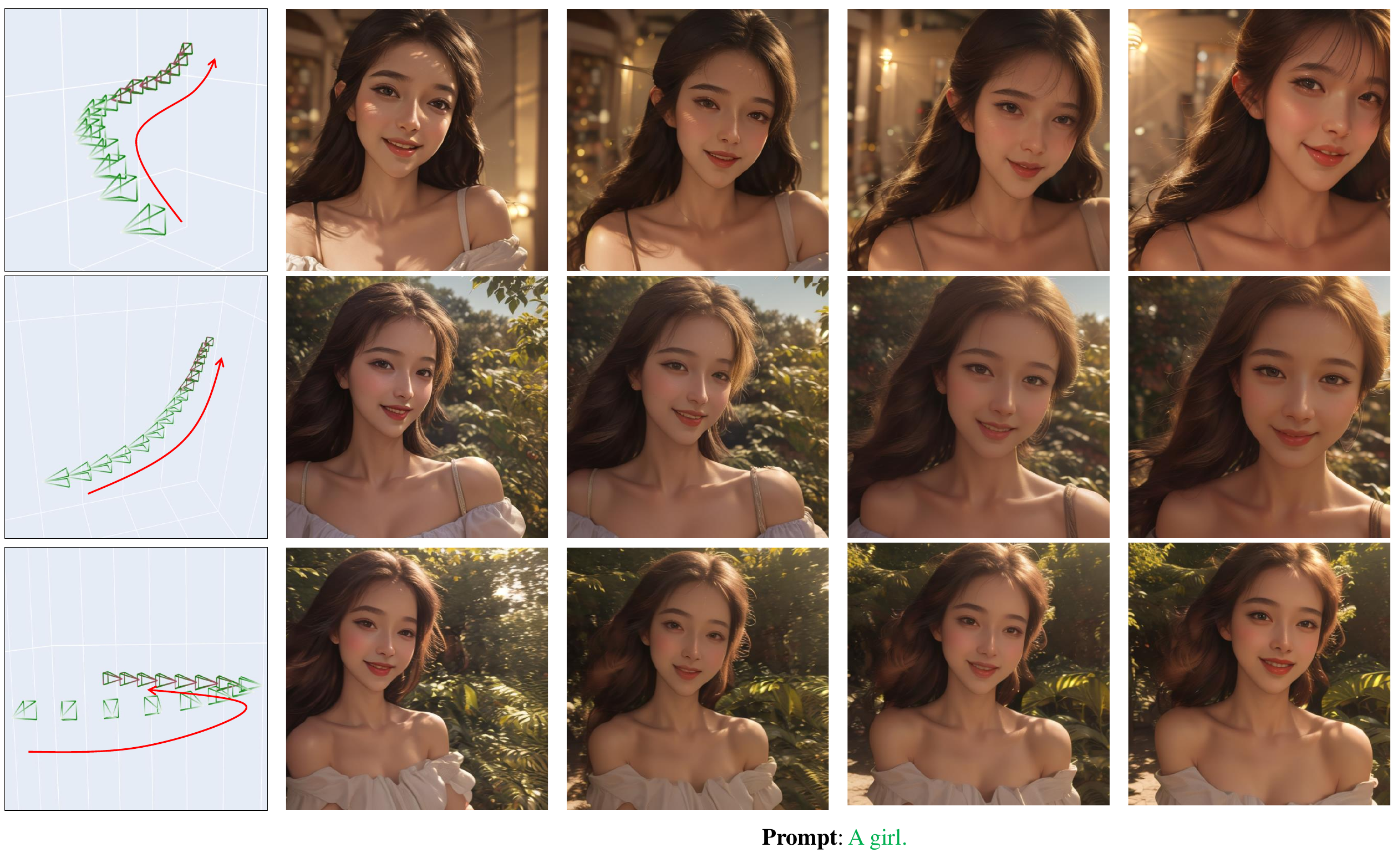} 
\vspace{-0.3cm}
\caption{Results of complex camera motion control deployed on AnimateDiff~\cite{guo2023animatediff}}
\label{fig:supp_animate_complex} 
\end{figure*}

\begin{figure*}[t]
\hsize=\textwidth
\centering
\includegraphics[width=0.64\textwidth]{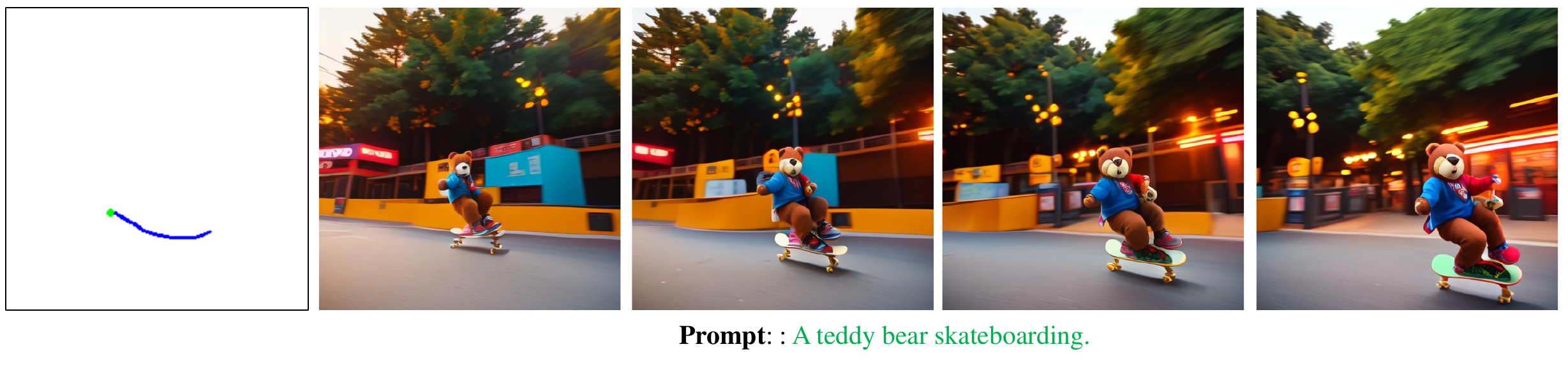} 
\vspace{-0.3cm}
\caption{Results of object motion control deployed on AnimateDiff~\cite{guo2023animatediff}. }
\label{fig:supp_animate_traj} 
\end{figure*}

\clearpage

\noindent The supplementary materials provide additional results achieved with our proposed MotionCtrl, along with in-depth analyses. \textbf{For a more visual understanding, we strongly recommend readers visit our project page for the video results}. The structure of the supplementary materials is as follows:
\begin{itemize}
    \item \textcolor{black}{Details of training data construction. (Section~\ref{sec:training_data})}
    \item \textcolor{black}{Details of evaluation datasets. (Section~\ref{sec:testset})}
    \item \textcolor{black}{More quantitative and qualitative results. (Section~\ref{sec:more_qualitative_and_qualitative_resutls})}
    \item More Results of MotionCtrl when extended to AnimateDiff~\cite{guo2023animatediff} framework. (Section~\ref{sec:animatediff})
    \item More discussions about previous related works. (Section~\ref{sec:related_works})
\end{itemize}

\section{Details of Training Data Construction}
\label{sec:training_data}

\textcolor{black}{
\noindent\textbf{Augmented-RealEstate10K.}
The camera motion control module (CMCM) in MotionCtrl is trained with data augmented from RealEstate10K~\cite{zhou2018stereo}. RealEstate10K originally contains videos with annotations of camera poses. To adapt it to our MotionCtrl, we further synthesize captions for each video with Blip2~\cite{blip2}, an image captioning algorithm. Specifically, we extract frames at specific intervals—the first, quarter, half, three-quarters, and final frames of a video. We then use Blip2 to predict their captions. These captions are concatenated to form a comprehensive description for each video clip. With these captions in place, we train the CMCM on RealEstate10K, enabling effective camera motion control in video generation models such as LVDM~\cite{he2022lvdm}.
}

\textcolor{black}{
\noindent\textbf{Augmented-WebVid.}
The object motion control module (OMCM) in MotionCtrl is trained with data augmented from WebVid~\cite{bain2021frozen}. WebVid is a large-scale video dataset equipped with captions and commonly used in the T2V generation task. To adapt it to our MotionCtrl, we further synthesize the object movement trajectories for the videos in WebVid with ParticleSfM~\cite{zhao2022particlesfm}. Although ParticleSfM is a structure-from-motion system primarily, it incorporates a trajectory-based motion segmentation module utilized for filtering out dynamic trajectories that affect the production of camera trajectories in a dynamic scene. The dynamic trajectories attained by the motion segmentation module exactly fulfill the requirements of our MotionCtrl and we employ this module to synthesize moving object trajectories required by our MotionCtrl. However, despite its effectiveness, ParticleSfM is not time-efficient, requiring approximately 2 minutes to process a 32-frame video. To mitigate the issue of time efficiency, we randomly select 32 frames from each WebVid video, with a frame skip interval $s \in [1, 16]$, to synthesize the object movement trajectories. This approach yields a total of 243,000 video clips that fulfill the training requirements for the OMCM.
}

\section{Details of Evaluation Datasets}
\label{sec:testset}

In this paper, we construct two evaluation datasets to independently evaluate the efficacy of our proposed MotionCtrl on camera and object motion control, respectively. 

\begin{figure*}[ht]
\hsize=\textwidth
\centering

\includegraphics[width=0.7\textwidth]{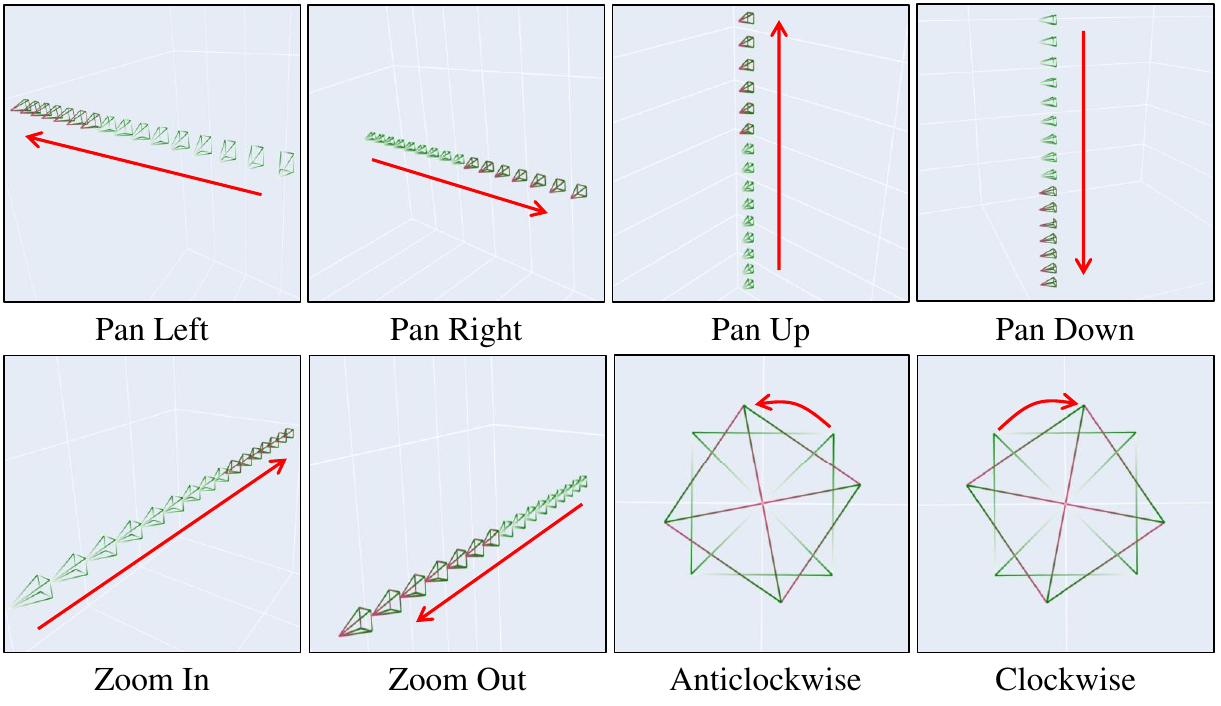} \\
\textbf{(a) 8 Basic Camera Poses} \\
\includegraphics[width=0.7\textwidth]{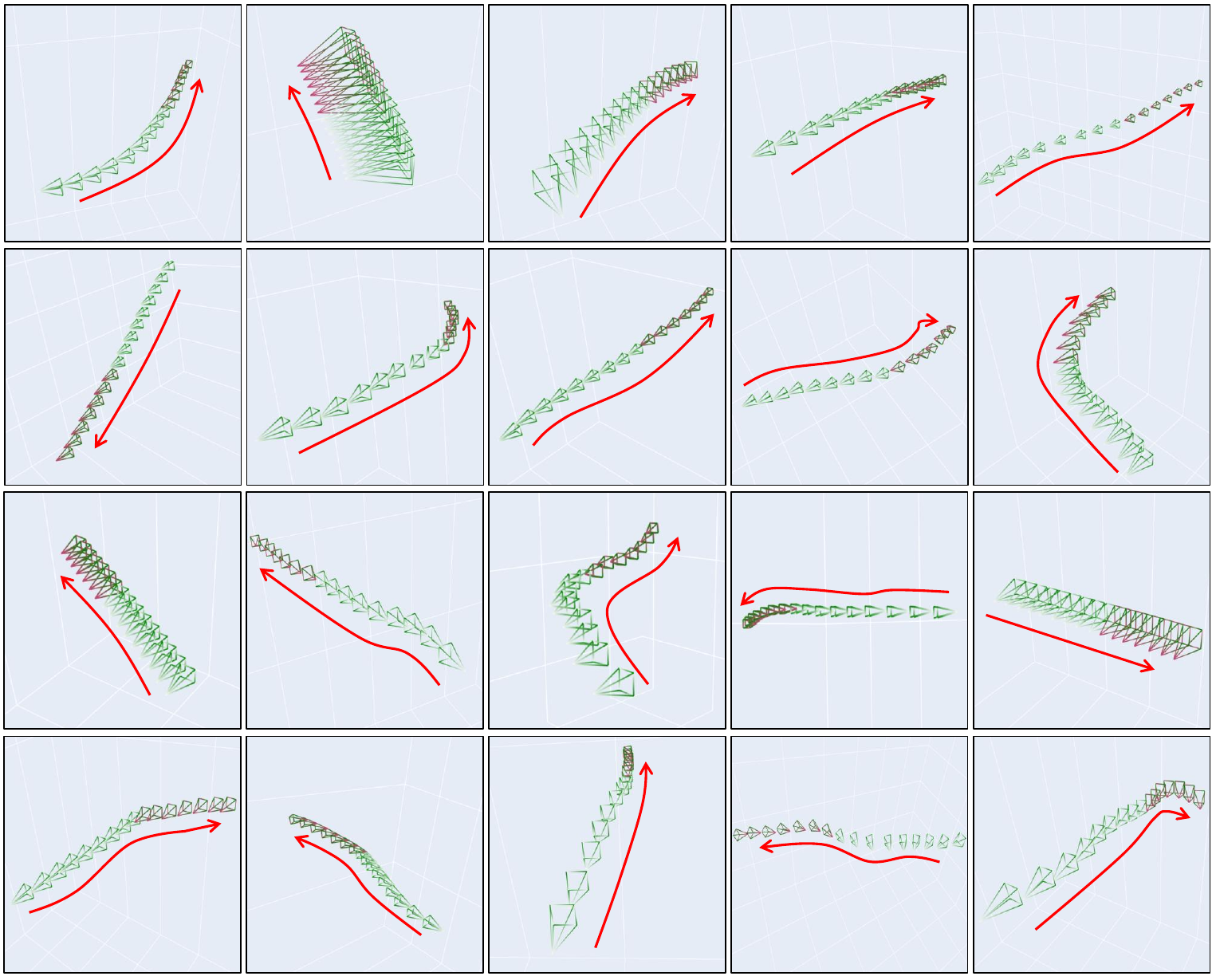} \\
\textbf{(b) 20 Relatively Complex Camera Poses from RealEstate10K Testset}
\caption{The \textbf{Camera Motion Control Evaluation Dataset} consists of 8 basic camera poses and 20 relatively complex camera poses, with the relatively complex poses being derived from the test set of RealEstate10K. This dataset is utilized to quantitatively assess the effectiveness of our proposed MotionCtrl in controlling a wide range of diverse camera motions in videos generated.}
\label{fig:cameraposes} 
\end{figure*}

\textcolor{black}{
\noindent\textbf{Camera Motion Control Evaluation Dataset.} This dataset contains a total of \textbf{407} samples covering two types of camera poses: 
\begin{enumerate}
    \item 80 ($8 \times 10$) samples constructed with 8 basic camera pose sequences (pan left, pan right, pan up, pan down, zoom in, zoom out, anticlockwise rotation, and clockwise rotation) and 10 prompts.
    \item 200 ($20 \times 10$) samples constructed with 20 relatively complex camera pose sequences randomly selected from the test set of RealEstate10K~\cite{zhou2018stereo} and 10 prompts.
    \item 100 samples constructed with 100 relatively complex camera poses of WebVid~\cite{bain2021frozen} synthesized with ParticleSfM~\cite{zhao2022particlesfm} and 100 prompts from VBench~\cite{huang2023vbench}.
    \item 27 samples constructed with 27 relatively complex camera poses of HD-VILA~\cite{xue2022advancing} synthesized with ParticleSfM and 27 prompts from VBench~\cite{huang2023vbench}.
\end{enumerate}
}
To provide an intuitive perception of the camera movement, we visualized the 8 basic camera poses and 20 relatively complex camera poses from RealEstate10K~\cite{zhou2018stereo} in Fig.~\ref{fig:cameraposes}. As described in the manuscript, the term "complex camera poses" as used in this work denotes camera movements beyond the basic camera poses list, encompassing camera turning and self-rotation within the same camera pose.

\begin{figure*}[ht]
\hsize=\textwidth
\centering
\includegraphics[width=0.7\textwidth]{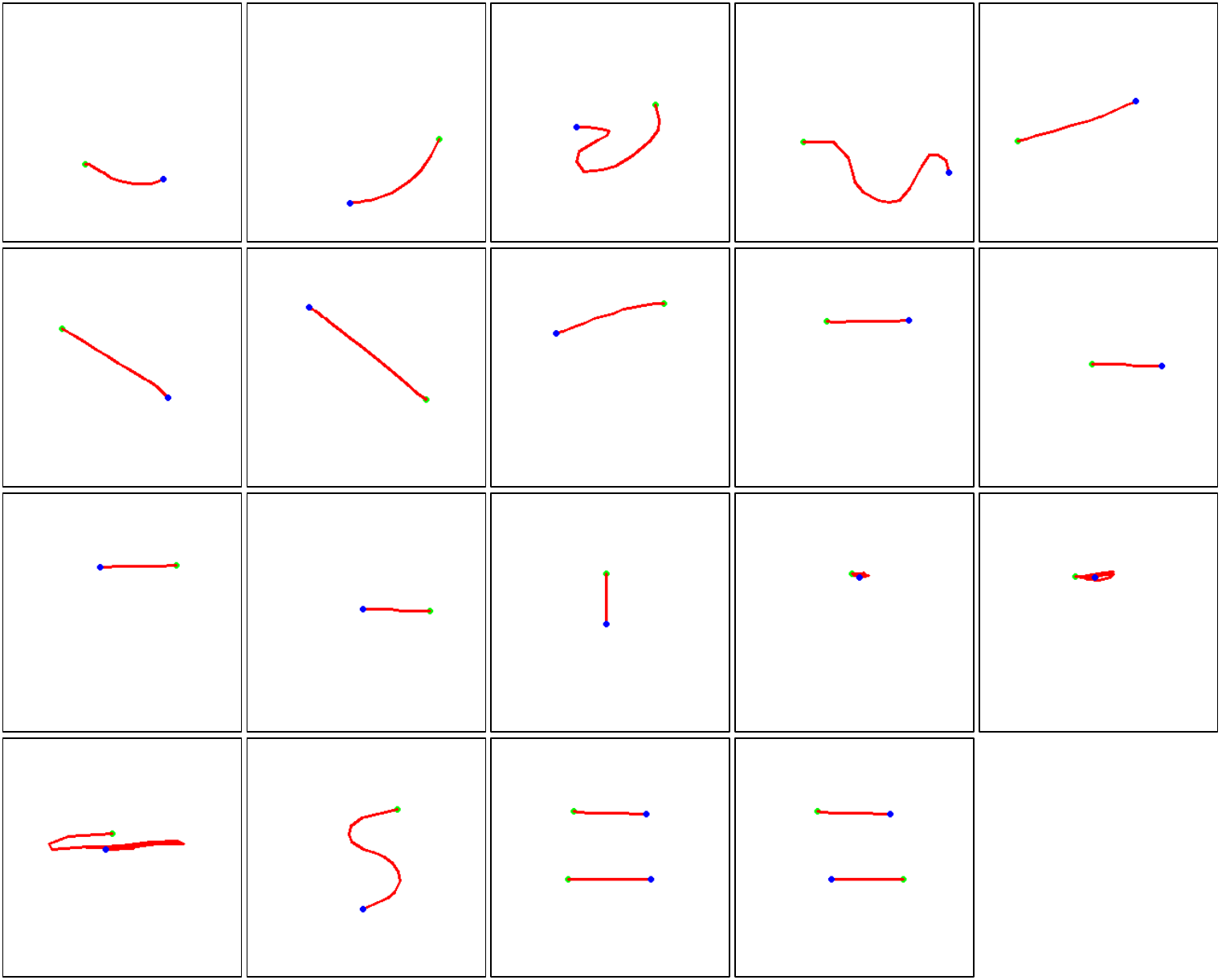} 
\caption{The \textbf{Object Motion Control Evaluation Dataset} encompasses 19 trajectories, where the \textcolor{green}{green} and \textcolor{blue}{blue} points respectively represent the starting and ending points of each trajectory. This dataset is used to quantitatively evaluate the effectiveness of the proposed MotionCtrl in controlling object movements in videos generated.}
\label{fig:trajs} 
\end{figure*}

\textcolor{black}{
\noindent\textbf{Object Motion Control Evaluation Dataset.} This evaluation dataset contains a total of 283 samples constructed with 74 diverse trajectories and 77 prompts. It should be noted that to verify the effectiveness of MotionCtrl in object motion control, our evaluation dataset pairs one trajectory with several different prompts or one prompt with several different trajectories. To provide an intuitive perception of the handcrafted trajectories, 19 trajectories adopted in the evaluation dataset are depicted in Fig.~\ref{fig:trajs}.
}

These evaluation datasets will be released.

\textbf{Please note that the evaluation datasets we have constructed are primarily used for quantitatively assessing the performance of our proposed MotionCtrl in both camera and object motion control in video generation. Our MotionCtrl is capable of handling a wider variety of camera poses and trajectories that are not included in the evaluation datasets.}

\section{More Quantitative and Qualitative Results}
\label{sec:more_qualitative_and_qualitative_resutls}

\subsection{More Quantitative Results}

\textcolor{black}{
\noindent\textbf{More Quantitative Comparisons on Relatively Complex Camera Motion Control.} In the manuscript, the quantitative results of relatively complex camera poses are statistics from all the complex camera poses sourced from RealEstate10K~\cite{zhou2018stereo}, WebVid~\cite{bain2021frozen}, and HD-VILA~\cite{xue2022advancing}. The statistical results for each dataset are presented in Table~\ref{tab:detail_complex}, demonstrating that our MotionCtrl outperforms VideoComposer~\cite{wang2023videocomposer} in both the camera poses extracted from RealEstate10K and those synthesized with ParticleSfM~\cite{zhao2022particlesfm} (camera poses of WebVid~\cite{bain2021frozen} and HD-VILA~\cite{xue2022advancing}) in terms of camera motion control, text similarity, and generated quality.
}

\begin{table*}[th]
  \caption{\textcolor{black}{\textbf{Quantitative Comparisons} with VideoComposer~\cite{wang2023videocomposer}. Our \textbf{MotionCtrl} performs better in all three sets of relatively complex camera poses from RealEstate10K~\cite{zhou2018stereo}, WebVid~\cite{bain2021frozen}, and HD-VILA~\cite{xue2022advancing}.}}
  \centering
  \begin{tabular}{c|cc|cc|cc}
    \toprule
    & \multicolumn{2}{|c|}{RealEstate10K} & \multicolumn{2}{c|}{WebVid} & \multicolumn{2}{c}{HD-VILA} \\ 
    Method & VideoComposer & \textbf{MotionCtrl} & VideoComposer & \textbf{MotionCtrl} & VideoComposer & \textbf{MotionCtrl} \\
    \midrule
    \textbf{CamMC~$\downarrow$}  & 0.1073  & \textbf{0.0840} & 0.0702  & \textbf{0.0589} & 0.0953  & \textbf{0.0499}\\
    \textbf{CLIPSIM~$\uparrow$}  & 0.2219  & \textbf{0.2324} & 0.2147  & \textbf{0.2268} & 0.2429  & \textbf{0.2473}\\
    \textbf{FID~$\downarrow$}    & 134.97  & \textbf{130.29} & 106.89  & \textbf{102.13} & 190.54  & \textbf{159.52}\\
    \textbf{FVD~$\downarrow$}    & 1045.82 & \textbf{934.37} & 733.09 & \textbf{612.84} & 1709.59 & \textbf{1129.40}\\
    \bottomrule
  \end{tabular}
  \label{tab:detail_complex}
\end{table*}

\noindent\textbf{User Study.}\textcolor{black}{
For a more comprehensive evaluation, we conduct a user study involving 34 participants to assess the results of VideoComposer~\cite{wang2023videocomposer} and MotionCtrl. The results were generated using object trajectories and relatively complex camera poses covering datasets from RealEstate10K~\cite{zhou2018stereo}, WebVid~\cite{bain2021frozen}, and HD-VILA~\cite{xue2022advancing}. The assessment included criteria such as Video Quality, Text Similarity, and Motion Similarity. Participants are also asked to express their overall preference for each compared pair. The statistical results in Table~\ref{tab:user_study} demonstrate that over 90 percent of participants preferred our results in all assessment aspects. Although VideoComposer exhibited good performance in motion control conditioned on motion vectors, its generated videos often appeared unnatural and strange due to the object shapes captured by the motion vectors from the reference video. Consequently, users showed a stronger preference for our relatively natural results.
}

\begin{table}[th]
  \caption{\textcolor{black}{\textbf{User Study.} Compared to the results generated with VideoComposer~\cite{wang2023videocomposer}, our \textbf{MotionCtrl} achieved more preference in all assessment aspect.}}
  \centering
  \begin{tabular}{c|cc}
    \toprule
    Method                                  & VideoComposer & \textbf{MotionCtrl} \\
    \midrule
    \textbf{Quality~$\uparrow$}             & 0.0628  & \textbf{0.9372} \\
    \textbf{TextSimilarity~$\uparrow$}      & 0.0772  & \textbf{0.9228} \\
    \textbf{MotionSimilarity~$\uparrow$}    & 0.086   & \textbf{0.9140} \\
    \textbf{OverallPreference~$\uparrow$}   & 0.0739  & \textbf{0.9261} \\
    \bottomrule
  \end{tabular}
  \label{tab:user_study}
\end{table}

\subsection{More Qualitative Results}
\begin{figure*}[ht]
\hsize=\textwidth
\centering
\includegraphics[width=0.80\textwidth]{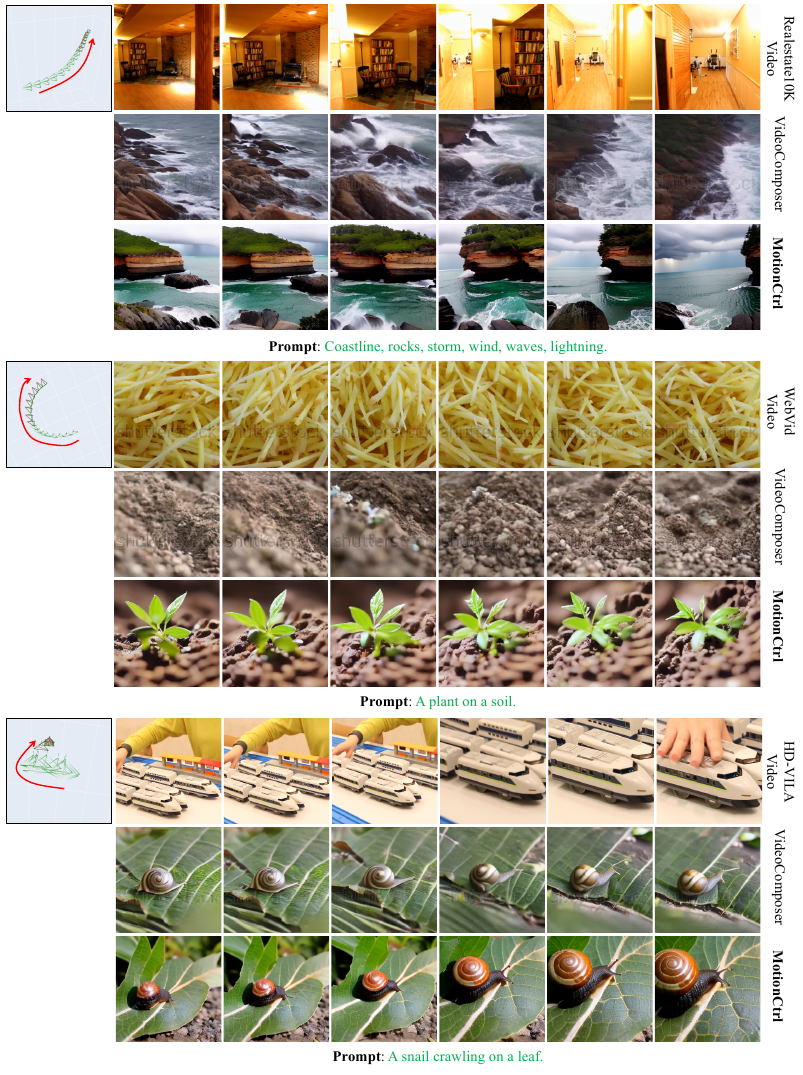} 
\caption{\textcolor{black}{\textbf{More Qualitative Comparisons with VideoComposer~\cite{wang2023videocomposer} on Camera Motion Control.} The generated videos of MotionCtrl can better follow the camera poses, whether from RealEstate10K~\cite{zhou2018stereo} or those synthesized with ParticleSfM~\cite{zhao2022particlesfm} on videos of WebVid~\cite{bain2021frozen} and HD-VILA~\cite{xue2022advancing}. Moreover, the results achieved with MotionCtrl exhibit higher quality.}}

\label{fig:supp_more_camera} 
\end{figure*}

\begin{figure*}[ht]
\hsize=\textwidth
\centering
\includegraphics[width=0.9\textwidth]{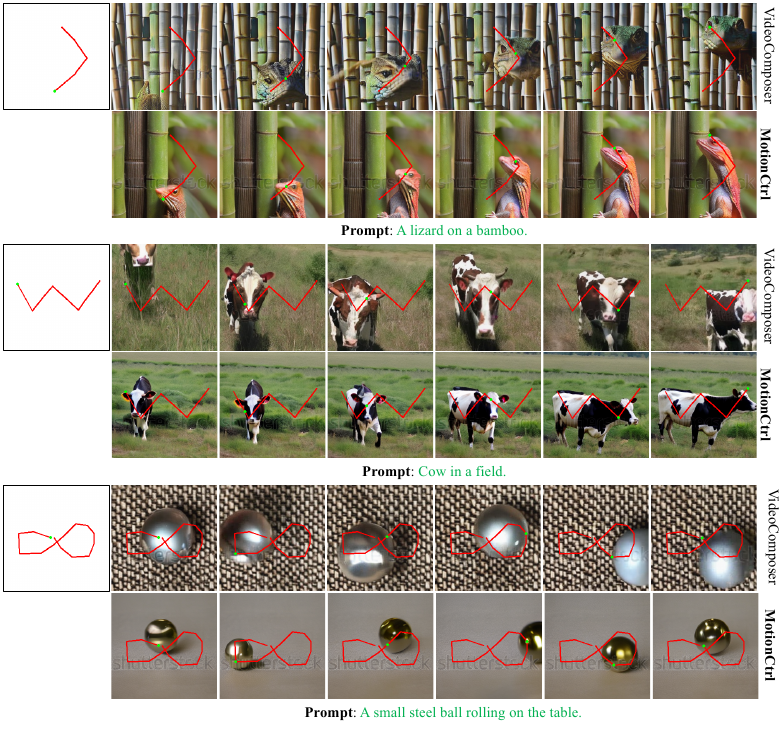} 
\caption{\textcolor{black}{\textbf{More Qualitative Comparisons with VideoComposer~\cite{wang2023videocomposer} on Object Motion Control.} The generated videos of MotionCtrl exhibit a superior ability to follow the trajectories in each frame and achieve higher overall quality.}}
\label{fig:supp_more_traj} 
\end{figure*}

\textcolor{black}{
\noindent\textbf{More Qualitative Comparisons with VideoComposer.}
We present additional qualitative results comparing VideoComposer~\cite{wang2023videocomposer} and our proposed MotionCtrl on relatively complex camera and object trajectories in Fig.~\ref{fig:supp_more_camera} and Fig.~\ref{fig:supp_more_traj}, respectively. These results suggest that MotionCtrl outperforms VideoComposer in both camera and object motion control in generated videos. Furthermore, MotionCtrl's generated videos exhibit higher quality and its generated content is better aligned with the prompts.
}

\noindent\textbf{More of MotionCtrl.}
\label{sec:more_results}
In this section, we present additional results of MotionCtrl, focusing on camera motion control, object motion control, and combined motion control. \textbf{Notably, all results are obtained using the same trained MotionCtrl model, without the need for extra fine-tuning for different camera poses or trajectories.}

\begin{figure*}[ht]
\hsize=\textwidth
\centering
\includegraphics[width=0.85\textwidth]{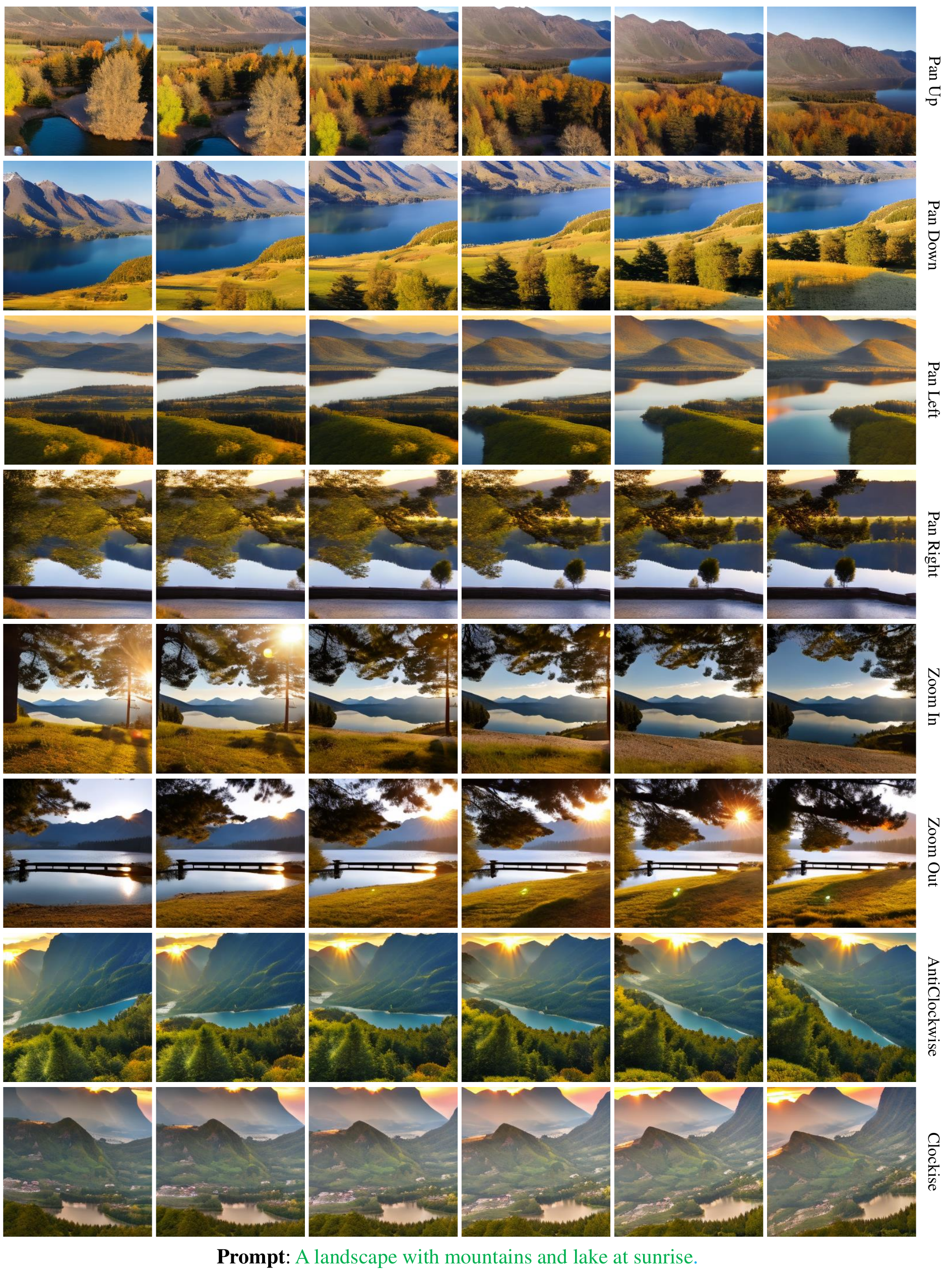} 
\caption{The results of our proposed MotionCtrl deployed on LVDM~\cite{he2022lvdm}, guided by 8 basic camera poses: pan up, pan down, pan left, pan right, zoom in, zoom out, anticlockwise rotation, and clockwise rotation (The visualization of these camera poses can be seen in Fig.~\ref{fig:cameraposes} (a)). \textbf{It's important to note that all results are achieved using the same MotionCtrl model, without the need for extra fine-tuning for different camera poses.}}
\label{fig:supp_results_8_camera_poses} 
\end{figure*}

Specifically, Fig.~\ref{fig:supp_results_8_camera_poses} illustrates the outcomes of camera motion control of MotionCtrl guided by 8 basic camera poses, including pan up, pan down, pan left, pan right, zoom in, zoom out, anticlockwise rotation, and clockwise rotation. These poses are visualized in Fig.~\ref{fig:cameraposes} (a). This demonstrates the capability of our MotionCtrl model to integrate multiple basic camera motion controls in a unified model, contrasting with the AnimateDiff model~\cite{guo2023animatediff} which requires a distinct LoRA model~\cite{lora} for each camera motion.

\begin{figure*}[ht]
\hsize=\textwidth
\centering
\includegraphics[width=0.9\textwidth]{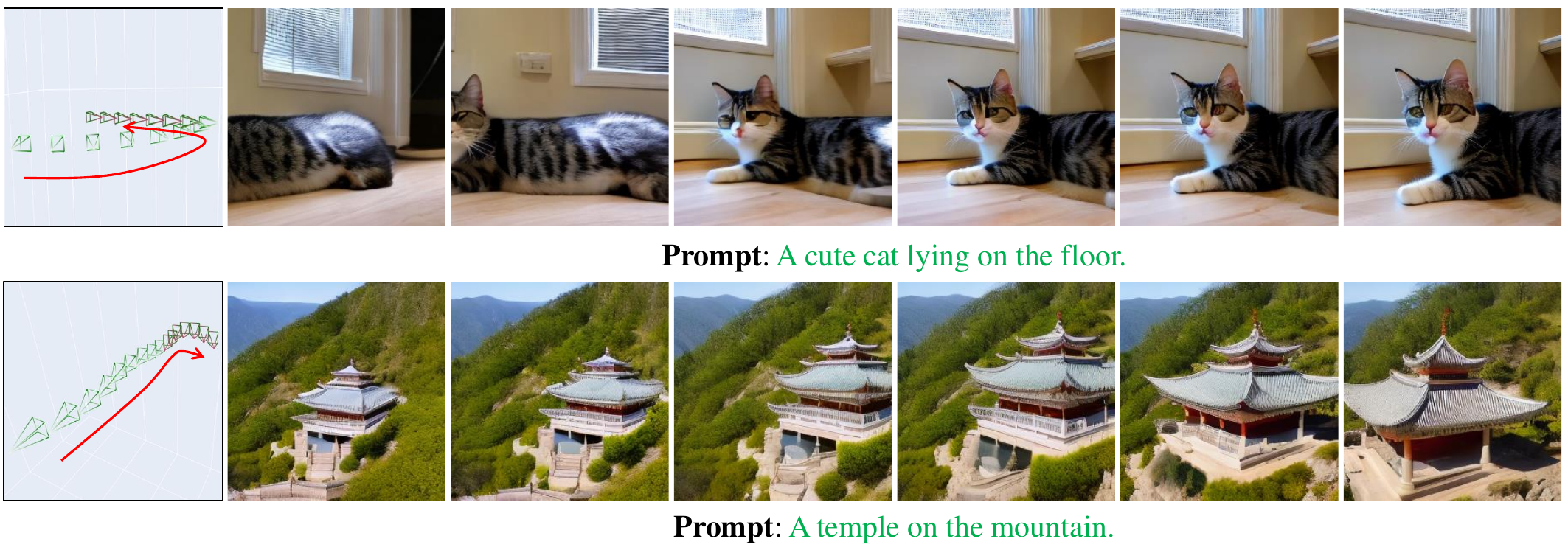} 
\caption{The results of our proposed MotionCtrl deployed on LVDM~\cite{he2022lvdm}, guided by relatively complex camera poses. \textbf{Unlike basic camera poses, which only involve simple directional movements, these complex camera poses incorporate elements of camera turning or self-rotation within the same camera pose sequence.} The camera motion in the generated videos closely follows the guided camera poses, while the generated content aligns with the text prompts.}
\label{fig:supp_results_complex_camera_poses} 
\end{figure*}

Fig.~\ref{fig:supp_results_complex_camera_poses} showcases the results of camera motion control using MotionCtrl, which is guided by relatively complex camera poses. \textbf{These complex camera poses are distinct from basic camera poses, as they include elements of camera turning or self-rotation within the same camera pose sequence}. The results demonstrate that, given a sequence of camera poses, our MotionCtrl can generate natural videos. The content of these videos aligns with the text prompts, and the camera motion corresponds to the provided complex camera poses.

\begin{figure*}[ht]
\hsize=\textwidth
\centering
\includegraphics[width=0.9\textwidth]{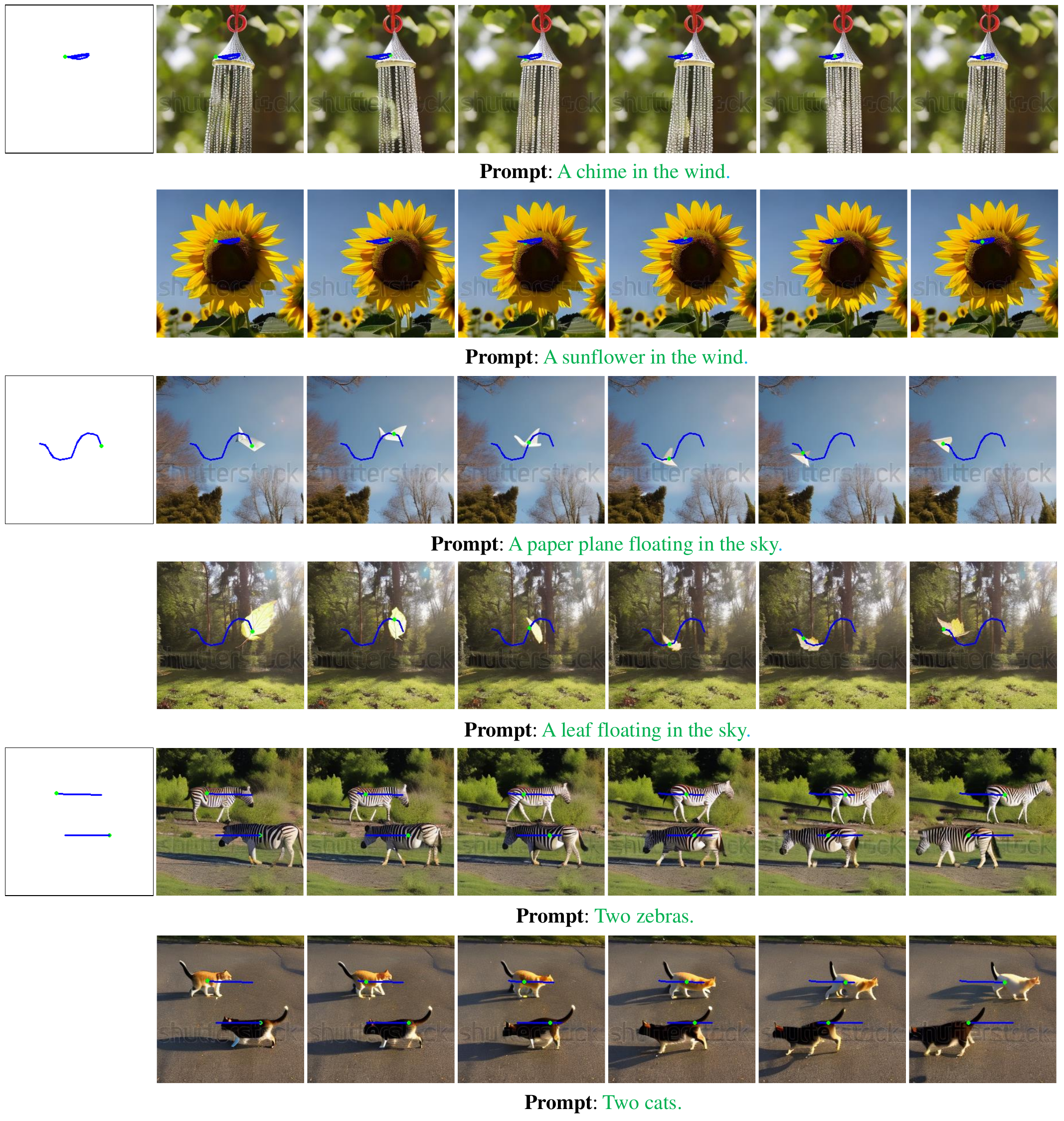} 
\caption{The result of our proposed MotionCtrl deployed on LVDM~\cite{he2022lvdm}, guided with trajectories. The \textcolor{green}{green points} in the trajectories indicate the starting points. Given the same trajectories, our model can generate different objects in accordance with the text prompts, maintaining the same object motion. When multiple trajectories are present in the same video, our model is capable of simultaneously controlling the motion of different objects within the same generated video.}
\label{fig:supp_results_trajs} 
\end{figure*}

Fig.~\ref{fig:supp_results_trajs} presents the results of object motion control using MotionCtrl, guided by specific trajectories. When given the same trajectories and different text prompts, MotionCtrl can generate videos featuring different objects, but with identical object motion. 

\begin{figure*}[ht]
\hsize=\textwidth
\centering
\includegraphics[width=0.9\textwidth]{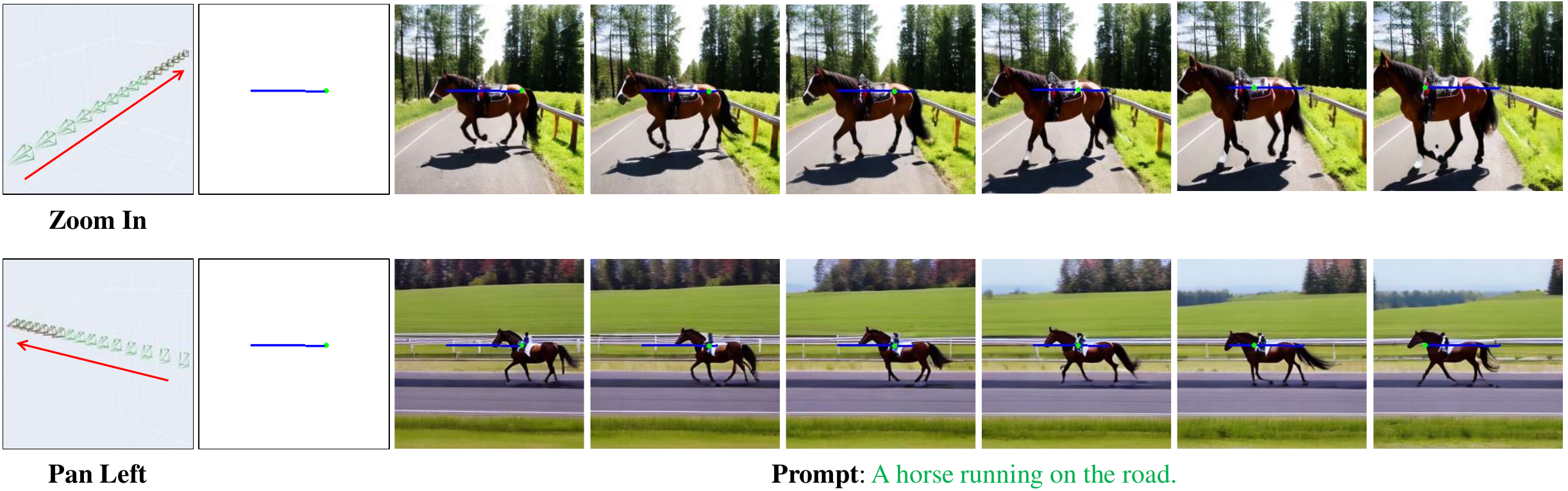} 
\caption{The result of combining camera motion and object motion control of MotionCtrl deployed on LVDM~\cite{he2022lvdm}. With the same trajectory but different camera poses, the horse in the generated videos has a different performance.}
\label{fig:supp_results_combine} 
\end{figure*}

Fig.~\ref{fig:supp_results_combine} provides the results of combining both the camera motion control and object motion control. With the same trajectory but different camera poses, the horse in the generated videos has a different performance. 

\begin{table*}[ht]
  \caption{\textbf{Differences between our proposed MotionCtrl and related works}. Unlike AnimateDiff~\cite{guo2023animatediff} (which refers to the motion control LoRA model provided by AnimateDiff), Tune-a-video~\cite{wu2022tune}, LAMP~\cite{wu2023lamp}, and MotionDirector~\cite{zhao2023motiondirector} that implement motion control by extracting motion from one or a series of template videos and require different models for different template videos, our proposed MotionCtrl uses a unified model. Besides, the motions learned by these methods are determined by the template video(s) and they do not distinguish between camera motion and object motion. On the other hand, although MotionDirector~\cite{zhao2023motiondirector} and VideoComposer~\cite{wang2023videocomposer} achieve motion control with a unified model guided by motion vectors and trajectories, respectively, they also do not distinguish between camera motion and object motion. In contrast, our proposed MotionCtrl, with a unified model, can independently and flexibly control the camera motion and object motion of the generated video, guided by camera poses and trajectories, respectively. }
  \label{tab:supp_related_work}
  \centering
  \resizebox{\linewidth}{!}{
  \begin{tabular}{c|cccc}
      \toprule
      Method                                       & Require Fine-tuning  & Motion sources & Distinguish Camera \& Object Motion  \\
      \midrule
      AnimateDiff~\cite{guo2023animatediff}        & \Checkmark   & template videos  & \XSolidBrush \\
      Tune-a-video~\cite{wu2022tune}               & \Checkmark   & template video  & \XSolidBrush \\
      LAMP~\cite{wu2023lamp}                       & \Checkmark   & template videos & \XSolidBrush \\
      MotionDirector~\cite{zhao2023motiondirector} & \Checkmark   & template videos & \XSolidBrush \\
      VideoComposer~\cite{wang2023videocomposer}   & \XSolidBrush & motion vectors   & \XSolidBrush \\
      DragNUWA~\cite{yin2023dragnuwa}              & \XSolidBrush & trajectories    & \XSolidBrush \\
      \textbf{MotionCtrl} (Ours)                   & \XSolidBrush & camera poses \& trajectories & \Checkmark \\
      \bottomrule
  \end{tabular}
  }
\end{table*}

\section{More Results of MotionCtrl Deployed on AimateDiff~\cite{guo2023animatediff}}
\label{sec:animatediff}

\begin{figure*}[ht]
\hsize=\textwidth
\centering
\includegraphics[width=0.6\textwidth]{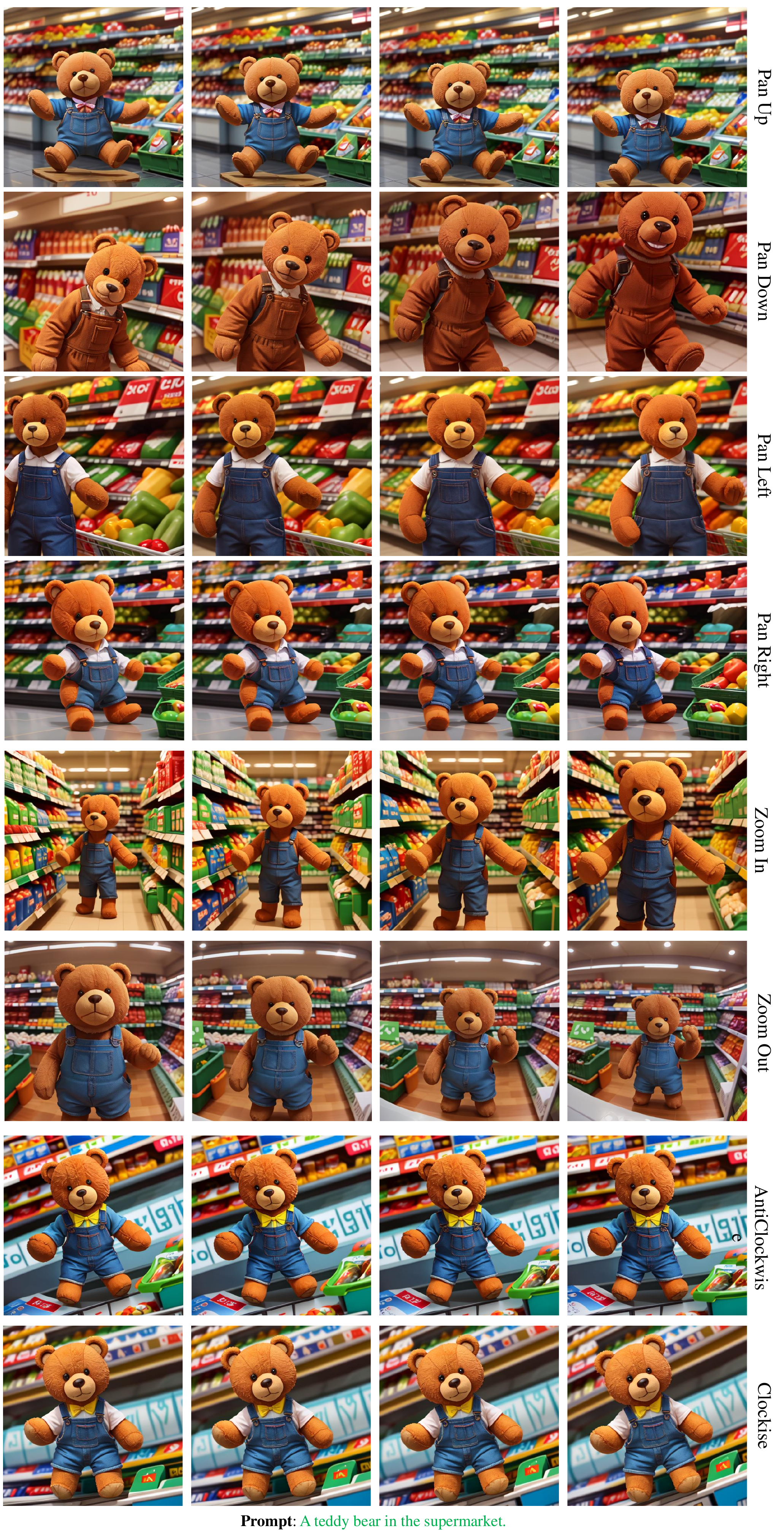} 
\caption{The camera motion control results of MotionCtrl deployed on AnimateDiff~\cite{guo2023animatediff}. They are guided with 8 basic camera poses.}
\label{fig:supp_animate_8} 
\end{figure*}

\begin{figure*}[ht]
\hsize=\textwidth
\centering
\includegraphics[width=0.6\textwidth]{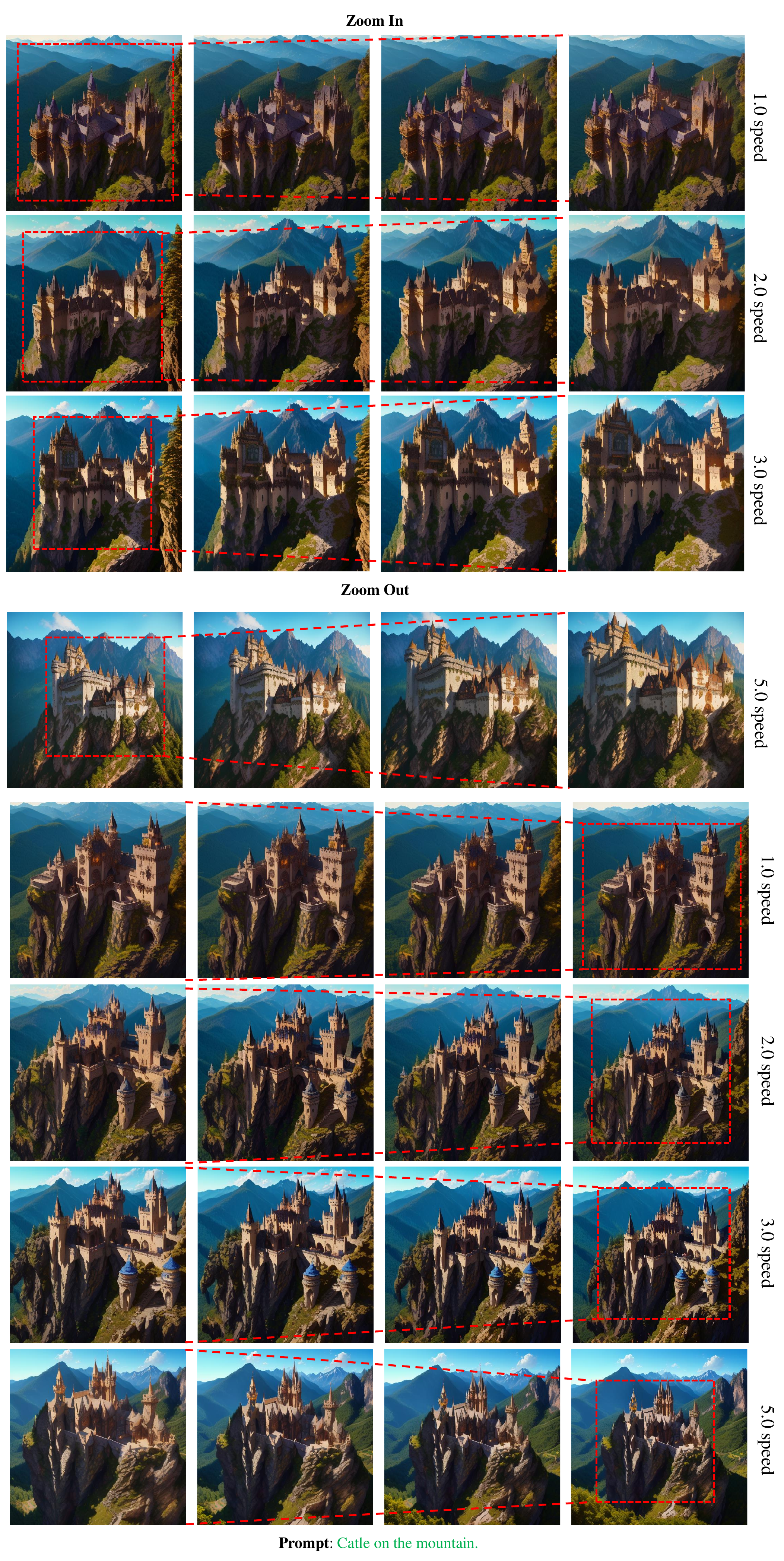} 
\caption{The camera motion control results of MotionCtrl deployed on AnimateDiff~\cite{guo2023animatediff}. Our MotionCtrl can not only control the camera motion of the generated videos but also their motion speed.}
\label{fig:supp_animate_speed} 
\end{figure*}

We also deploy our MotionCtrl on AnimateDiff~\cite{guo2023animatediff}. Therefore, we can control the motion of the video generated with our fine-tuned AnimateDiff cooperating with various LoRA~\cite{lora} models in the committee. Results of relatively complex camera motion control and object motion control are in the manuscripts and we provide the results of basic camera motion control here: Fig.~\ref{fig:supp_animate_8} and Fig.~\ref{fig:supp_animate_speed}. These results are generated with our MontionCtrl cooperating with different LoRA models provided by in CIVITAI~\cite{civitai}. They demonstrate that our the generalization of MotionCtrl that can be adapted to different video generation models.

\section{More Discussions about the Related Works}
\label{sec:related_works}

To further illustrate the advantages of our proposed MotionCtrl, we've conducted a comparative analysis with previous related works. The comparisons are detailed in Table~\ref{tab:supp_related_work}.
Models such as AnimateDiff\cite{guo2023animatediff} (refers to the motion control LoRA models provided by AnimateDiff), Tune-a-video\cite{wu2022tune}, LAMP\cite{wu2023lamp}, and MotionDirector\cite{zhao2023motiondirector} implement motion control by extracting motion from one or multiple template videos. This approach necessitates the training of distinct models for each template video or template video set. Moreover, the motions these methods learned are solely determined by the template video(s), and they fail to differentiate between camera motion and object motion.
Similarly, MotionDirector\cite{zhao2023motiondirector} and VideoComposer\cite{wang2023videocomposer}, despite achieving motion control with a unified model guided by motion vectors and trajectories respectively, do not distinguish between camera motion and object motion.
In contrast, our proposed MotionCtrl, utilizing a unified model, can independently and flexibly control a wide range of camera and object motions in the generated videos. This is achieved by guiding the model with camera poses and trajectories respectively, offering a more fine-grained control over the video generation process.

\end{document}